\documentclass[sigconf,nonacm]{acmart}
\AtBeginDocument{%
  }

\usepackage{ulem}

\usepackage[noabbrev]{cleveref}
\usepackage{booktabs}
\usepackage{multirow}
\usepackage{makecell}
\usepackage{xcolor}
\usepackage{colortbl}

\usepackage{listings} 
\usepackage{tcolorbox} 
\usepackage{caption}
\usepackage{subcaption}
\tcbuselibrary{breakable,skins,listings} 
\newtcolorbox{promptbox}[1][]{
  enhanced,
  breakable,
  colback=gray!5,
  colframe=gray!40,
  title=Prompt Template, 
  #1, 
  boxrule=0.5pt,
  left=2pt, right=2pt, top=2pt, bottom=2pt,
  before skip=6pt,
  after skip=6pt,
}

\usepackage{pifont}

\begin{document}

\title{FlipVQA: Scaling Multi-modal Instruction Tuning via Textbook-to-Knowledge Synthesis}

\author{Zhen Hao Wong}
\affiliation{%
  \institution{Peking University}
  \city{Beijing}
  \country{China}
}
\email{zhenhao1141@stu.pku.edu.cn}

\author{Jingwen Deng}
\affiliation{%
  \institution{Peking University}
  \city{Beijing}
  \country{China}}
\email{dengjingwen@stu.pku.edu.cn}

\author{Yuzhao Wang}
\affiliation{%
  \institution{Peking University}
  \city{Beijing}
  \country{China}
}
\email{yuzhaowang25@stu.pku.edu.cn}

\author{Wenkai Yu}
\affiliation{%
  \institution{Peking University}
  \city{Beijing}
  \country{China}
}
\email{ywk2022@stu.pku.edu.cn}

\author{Jihao Huang}
\affiliation{%
  \institution{Peking University}
  \city{Beijing}
  \country{China}
}
\email{2300017786@stu.pku.edu.cn}

\author{Runming He}
\affiliation{%
  \institution{Peking University}
  \city{Beijing}
  \country{China}
}
\email{hrm1165444624@stu.pku.edu.cn}

\author{Chengyu Shen}
\affiliation{%
  \institution{Peking University}
  \city{Beijing}
  \country{China}
}
\email{2501210669@stu.pku.edu.cn}

\author{Hao Liang}
\affiliation{%
  \institution{Peking University}
    \city{Beijing}
  \country{China}
}
\affiliation{%
  \institution{Zhongguancun Academy}
  \city{Beijing}
  \country{China}}
\email{hao.liang@stu.pku.edu.cn}

\author{Wentao Zhang}
\affiliation{%
  \institution{Peking University}
  \city{Beijing}
  \country{China}
}
\affiliation{%
  \institution{Zhongguancun Academy}
  \city{Beijing}
  \country{China}}

\email{wentao.zhang@stu.pku.edu.cn}


\begin{abstract}
Textbooks are among the richest repositories of human-verified reasoning knowledge, yet their complex layouts contain multi-column typesetting, cross-page question answer separation, and interleaved figures, make automated extraction of structured QA and VQA pairs extremely challenging. Existing alternatives either synthesize data from scratch, which lacks authentic problem contexts, or rely on costly expert annotation that cannot scale. We propose \textbf{FlipVQA-Miner}, an automated pipeline that resolves long-range logical dependencies and cross-page discontinuities in OCR-parsed documents, recovering coherent question--answer--figure associations even when answers reside in separate companion volumes. A subsequent multi-stage curation pipeline transforms these raw extractions into AI-ready supervision signals. Using FlipVQA-Miner, we construct \textbf{FlipVQA-83K}, comprising 83K QA and VQA pairs spanning 11 academic disciplines, at a \textbf{50$\times$} cost saving compared to manual annotation while maintaining high structural fidelity ($F_1 > 0.96$). Models fine-tuned on FlipVQA-83K demonstrate significantly improved reasoning ability and cross-domain generalization, establishing a scalable paradigm for human-knowledge-grounded data curation. Our
dataset and the complete data generating and curating methods
can be found in \url{https://github.com/OpenDCAI/DataFlow-VQA}.
\end{abstract}




\maketitle

\section{Introduction}
The reasoning capability of Large Language Models (LLMs) has become central to a wide range of applications, including complex problem solving, mathematical theorem proving, and multimodal understanding \cite{guo2025deepseek,gpt-5}. As modern models increasingly rely on reasoning ability, the availability of high-fidelity, reasoning-oriented multimodal data has emerged as a critical bottleneck.

Textbooks constitute one of the most valuable sources of such data: they provide expert-authored problems with verified solutions, grounded in authentic pedagogical contexts across diverse disciplines. Early educational multimodal QA benchmarks such as TQA and ScienceQA already suggest that textbook-style materials are highly valuable sources of reasoning supervision \cite{kembhavi2017tqa,lu2022scienceqa}. However, automatically extracting structured QA and VQA pairs from textbooks remains a challenging and largely underexplored problem. While modern OCR-based document parsing systems can reliably recover low-level layout elements, such as text blocks, equations, tables, and figures, the resulting outputs lack the semantic structure required for downstream learning\cite{smith2007tesseract,xu2020layoutlm,huang2022layoutlmv3,lee2023pix2struct,niu2025mineru2}. Questions and answers are often separated by dozens of pages, figures are detached from the problems that reference them, and multi-part exercises are fragmented across columns and page boundaries. Without resolving these long-range logical dependencies, the parsed content cannot be directly transformed into usable training data.

Existing alternatives to textbook-based data construction suffer from complementary limitations. \textbf{Synthetic data generation} pipelines can efficiently produce large volumes of structured samples , and self-instruct style expansion has greatly lowered the marginal cost of dataset creation \cite{wang2023self,tan2024large}; however, such data often lack authentic context, compositional depth, and pedagogical diversity, limiting their effectiveness for real-world reasoning. In contrast, \textbf{expert-curated educational datasets} better reflect realistic learning scenarios~\cite{kembhavi2017tqa,lu2022scienceqa}; however, their reliance on domain experts for annotation, verification, and quality control makes them difficult to scale, both in terms of cost and time, thereby constraining dataset size and diversity.

To address these challenges, we introduce \textbf{FlipVQA-Miner}, an automated pipeline that extracts high-quality QA and VQA pairs directly from OCR-parsed pedagogical documents. Unlike prior document parsing approaches that primarily focus on layout reconstruction, our framework introduces an Identifier-Guided Semantic Reconstruction mechanism \cite{huang2022layoutlmv3,zhao2024doclayout,lee2023pix2struct,niu2025mineru2,niu2025mineru2}. This design combines layout-aware parsing with LLM-assisted block grouping and logical alignment, enabling the recovery of coherent question-answer-figure associations even when answers are located far from their corresponding questions, such as at the end of a book or in companion solution manuals. 

Building upon this extraction framework, we further develop a multi-stage data curation pipeline that transforms raw extractions into AI-ready supervision signals through structured question reconstruction, answer verification, question-type classification, and quality filtering. By replacing traditional expert-driven annotation with automated reconstruction and curation, FlipVQA-Miner reduces dataset construction cost by over \textbf{50$\times$} while maintaining high data fidelity.

Using this system, we construct \textbf{FlipVQA-83K}, a large-scale reasoning dataset containing 83K QA and VQA instances spanning 11 academic disciplines. Extensive experiments demonstrate the effectiveness of the proposed data. Fine-tuning Qwen3-8B-Base on FlipVQA consistently improves reasoning performance across multiple subject benchmarks, surpassing the stronger instruction-tuned baseline Qwen3-8B-Instruct. In multimodal evaluation, the resulting model also outperforms Qwen3-VL-8B-Instruct on VQA benchmarks, highlighting the value of high-quality textbook-derived supervision.

\noindent\textbf{Contributions.}
\textbf{(1) Automated Pedagogical Data Mining.}
We introduce FlipVQA-Miner, an automated pipeline for mining human-authored educational materials, enabling scalable construction of high-fidelity QA and VQA data with minimal hallucination.

\textbf{(2) A Large-Scale Multi-Disciplinary Reasoning Dataset.}
We build FlipVQA-83K, a dataset consisting of 83K QA and VQA instances across 11 academic disciplines, derived directly from authentic pedagogical sources.

\textbf{(3) Cost-Efficient Data Construction.}
Compared with expert-driven annotation pipelines, FlipVQA-Miner reduces data construction cost by over \textbf{50$\times$}, significantly improving scalability without sacrificing data quality.

\textbf{(4) Strong Empirical Performance.}
Fine-tuning Qwen3-8B-Base on FlipVQA-83K yields consistent gains across multiple reasoning benchmarks, surpassing Qwen3-8B-Instruct and outperforming Qwen3-VL-8B-Instruct on VQA tasks.

\section{Related Work}

\subsection{Document Intelligence and Semantic Reconstruction.}
Document intelligence systems have transitioned from basic OCR to sophisticated layout-aware parsing. Traditional tools like Tesseract convert scanned pages into text but often discard crucial spatial metadata~\cite{smith2007tesseract}. Modern frameworks, including LayoutLMv3 , DocLayout-YOLO , Pix2Struct , and MinerU 2.5 , have achieved impressive accuracy in recovering structural elements such as tables, equations, and multi-column layouts\cite{xu2020layoutlm,huang2022layoutlmv3,kim2022donut,zhao2024doclayout,lee2023pix2struct,blecher2023nougat,niu2025mineru2}. However, these approaches are primarily optimized for document readability and information extraction rather than the synthesis of structured supervision signals, which do not address the semantic alignment required for training reasoning-oriented models, such as identifying question-answer boundaries or linking figures to dispersed textual context. 

\subsection{Multimodal Reasoning and Instruction Datasets.}
The evolution of Multimodal Large Language Models (MLLMs) has been significantly propelled by instruction-tuning datasets such as LLaVA-Instruct-150K, M3IT, ScienceQA, and TQA \cite{liu2023llava,li2023m3it,lu2022scienceqa, kembhavi2017tqa}. These datasets have established critical benchmarks for multimodal alignment and reasoning evaluation. However, existing corpora predominantly rely on synthetic generation via self-instruct pipelines or web-scale image-text pairing, which often introduces hallucination, stylistic homogeneity, and limited generalization. Furthermore, while benchmarks like MathVista address scientific reasoning~\cite{lumathvista,yue2024mmmu,he2024olympiadbench}, they are typically manually curated at a small scale or lack an automated pipeline to transform raw pedagogical materials into structurally coherent supervision. Specifically, existing datasets fail to preserve long-range, cross-page logical dependencies inherent in educational documents.

\section{Methodology}

\begin{figure*}
\centering
\vspace{-1em}
\includegraphics[width=\linewidth]{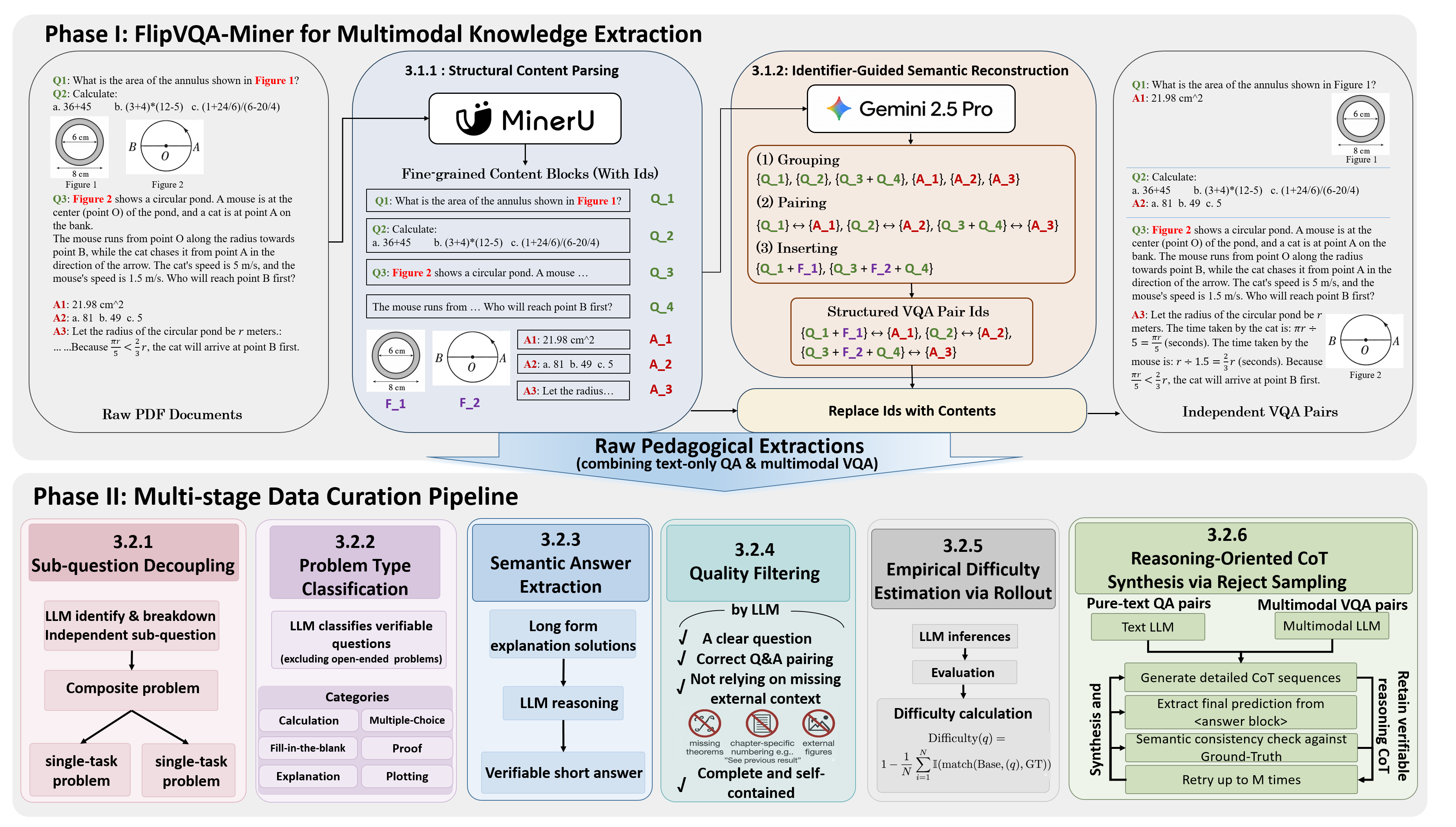}
\caption{Overview of FlipVQA: The pipeline consists of two primary stages: (1) FlipVQA-Miner, which extract structured QA and VQA from Textbooks; and (2) a Multi-stage Data Curation Pipeline, which curate the extracted data ready for training.}
\label{fig:overview}
\Description{}
\end{figure*}

In this section, we introduce \textbf{FlipVQA-Miner} for high-fidelity knowledge extraction, and \textbf{Multi-stage Data Curation Pipeline} to curate the extracted data ready for training.

\subsection{FlipVQA-Miner: Multimodal Knowledge Extraction}
The extraction phase employs a hybrid design that integrates layout-aware document parsing with LLM-based semantic reconstruction. Crucially, our approach is modality-agnostic, enabling the synthesis of both text-only Question-Answer (QA) pairs and multimodal Visual Question-Answer (VQA) pairs.

\subsubsection{Structural Content Parsing}
We first utilize MinerU 2.5~\citep{niu2025mineru2}, a powerful VLM-based OCR tool, to convert raw PDF files into a structured, machine-readable format. This step decomposes the document into fine-grained content blocks including text, equations, and embedded images while preserving their spatial metadata and layout information. By establishing this structural foundation, we provide a clean decomposition that reduces the complexity for subsequent semantic processing.

\subsubsection{Identifier-Guided Semantic Reconstruction (ISR)}
\label{sec:id_reconstruct}
To resolve long-range dependencies and semantic discontinuities, the LLM operates on unique block identifiers rather than raw content, substantially reducing the number of \textit{output tokens}. The system adaptively handles two extraction scenarios (see the prompt in \cref{app:id_reconstruct}): for \textbf{text-only QA extraction}, where sections are devoid of visual aids, the LLM focuses on \textbf{grouping} fragmented text and \textbf{pairing} questions with their corresponding answers across pages; for \textbf{multimodal VQA synthesis}, where visual evidence is present, the system additionally invokes an \textbf{inserting} operation, precisely placing relevant images into their associated textual context to form coherent VQA units.

\vspace{-1em}
\subsection{Multi-stage Data Curation Pipeline}

To transform raw pedagogical extractions into high-fidelity, ``AI-ready'' supervision signals, we implement a rigorous multi-stage curation pipeline. This process ensures that the resulting dataset is clean, verifiable, and self-contained, effectively bridging the gap between fragmented textbook content and structured instruction-tuning requirements. The LLM we use in this pipeline is GPT-5-mini~\citep{gpt-5}.

\subsubsection{Sub-question Decoupling}
\label{sec:subquestion_decompose}
Pedagogical materials frequently contain compound questions comprising multiple sub-questions that focus on one task (e.g. computing definite integral) but with different numbers. To ensure that each data entry poses a singular reasoning task, we employ a Sub-question Decoupling mechanism. Utilizing an LLM-based template generator, we identify and break down independent sub-problems while maintaining their relevant contexts. We carefully design the prompt so that context-sensitive sub-questions are not extracted. The prompt is provided in \cref{app:subquestion_decompose}.

\subsubsection{Problem Type Classification}
\label{sec:type_classification}
To facilitate objective evaluation, we perform Problem Type Classification on the extracted corpus. An LLM is prompted to categorize each problem into specific types, including \textit{Calculation}, \textit{Multiple-choice}, \textit{Fill-in-the-blank}, \textit{Proof}, \textit{Explanation} or \textit{Plotting}. In our dataset, we prioritize verifiable types (Calculation, Multiple-choice and Fill-in-the-blank) and exclude open-ended or proof-based problems that lacks objective verification metrics. The prompt is provided in \cref{app:type_classification}.

\subsubsection{Semantic Answer Extraction}
Textbook solutions often provide exhaustive, long-form explanations rather than concise, verifiable answers. We implement an Answer Extraction Operator that leverages LLM reasoning to derive canonical short answers from these narrative solutions, enabling automated evaluation. The prompt is provided in \cref{app:answer_extraction}.

\subsubsection{Quality Filtering}
\label{sec:data_quality}
To ensure each sample is pedagogically sound and self-contained, we apply a strict Quality Filtering protocol. An LLM is prompted to ensure that the question raises a clear problem (e.g., not stating a fact), and the question and solution are correctly paired. It also eliminate instances that rely on missing external context, such as references to prior theorems, chapter-specific numbering (e.g., ``See previous result''), or external figures. Each sample is thus verified to be complete and self-contained, providing all necessary information for the model to reach a solution independently. The prompt is provided in \cref{app:data_quality}.

\subsubsection{Empirical Difficulty Estimation via Rollout}
We quantify problem difficulty using an Empirical Rollout Evaluation ($N=32$). The difficulty coefficient is defined as $1 - \text{Avg}@N$ accuracy:
$$Difficulty(q) = 1 - \frac{1}{N} \sum_{i=1}^{N} \mathbb{I}(\text{match}(\text{Base}_i(q), \text{GT})),$$
where $\text{Base}$ represents the base model before fine-tuning, $\text{GT}$ is the groundtruth answer.
This enables the selection of challenging yet solvable samples, maintaining a reasonable difficulty distribution for training.

\subsubsection{Reasoning-Oriented CoT Synthesis via Reject Sampling}
\label{sec:cot_synthesis}
While textbooks provide high-quality problems and answers, their intermediate reasoning steps are often presented in a format designed for human pedagogy rather than machine supervision. To bridge this gap and endow models with explicit logical trajectories, we implement a Reject Sampling-based CoT Synthesis pipeline. Depending on the input modality, we utilize different reasoning-specialized models to generate detailed Chain-of-Thought (CoT) sequences ($\mathcal{T}$): specifically, Qwen3-235B-A22B-Thinking is employed for pure-text QA pairs, while Qwen3-VL-235B-A22B-Thinking is utilized for VQA pairs to ensure precise visual grounding. To ensure the faithfulness and logical integrity of the synthesized reasoning, we adopt a \textit{multi-round verification protocol}: for each generated candidate, we extract the final prediction from the \texttt{<answer>} block and perform a semantic consistency check against the ground-truth answer ($\text{GT}$). Only those trajectories that successfully lead to the correct result are retained:
$$ \mathcal{S} = \{ (q, \mathcal{T}) \mid \text{Verify}(\text{Extract}(\mathcal{T}), \text{GT}) = \text{True} \} $$
For instances where the initial generation fails, the pipeline performs up to $M$ ($M=5$ in our work) retries to maximize data yield without compromising quality. This iterative rejection and refinement process ensures that the resulting FlipVQA-83K dataset provides not only correct answers but also robust, verifiable reasoning chains for multimodal instruction tuning.

\section{FlipVQA-83K Dataset}
\begin{table*}[ht]
\centering
\resizebox{\textwidth}{!}{
\begin{tabular}{cccccccc}
\toprule
\textbf{Document} & \textbf{Type} & \textbf{Layout} & \textbf{Modality} & \textbf{\#Samples} & \textbf{Precision} & \textbf{Recall} & \textbf{F1} \\
\midrule
\multirow{2}{*}{\makecell[c]{Solutions Manual for Complex Analysis}} 
& \multirow{2}{*}{Interleaved} & \multirow{2}{*}{Single-column} & Text & 390 & 0.9797 & 0.9897 & 0.9847 \\
&  &  & Vision & 294 & 1.0000 & 0.9456 & 0.9720 \\
\midrule
\multirow{2}{*}{\makecell[c]{Textbook for Abstract Algebra}} 
& \multirow{2}{*}{Long-distance} & \multirow{2}{*}{Single-column} & Text & 1281 & 0.9968 & 0.9766 & 0.9866 \\
&  &  & Vision & 27 & 1.0000 & 0.9259 & 0.9615 \\
\midrule
\multirow{2}{*}{\makecell[c]{Exercise Book of Chinese Middle School Math }} 
& \multirow{2}{*}{Long-distance} & \multirow{2}{*}{Multi-column} & Text & 445 & 0.9911 & 0.9955 & 0.9933 \\
&  &  & Vision & 266 & 1.0000 & 0.9774 & 0.9886 \\
\bottomrule
\end{tabular}
}
\caption{Extraction performance of FlipVQA-Miner on different structural pattern documents. \#Samples indicates the number of QA pairs for text and the number of images for vision. Representative samples and visual layout demonstrations for each document are
provided in Appendix~\ref{app:case-study}.}
\label{tab:extraction-results}

\end{table*}
\subsection{Data Collection}
To construct a high-fidelity reasoning benchmark, we curate a corpus of 544 college-level educational PDF documents, including expert-authored textbooks and exercise sets\footnote{All documents are sourced from open-access repositories under permissive licenses (e.g., CC-BY 4.0 or Public Domain), ensuring ethical reuse and benchmark transparency.}. The collection spans 11 academic disciplines, primarily in STEM domains where problems typically involve rigorous and verifiable reasoning processes. The selected documents exhibit complex layouts, such as multi-column structures and cross-page dependencies, posing realistic challenges for multimodal understanding. 

\begin{figure}
\centering
\includegraphics[width=0.98\linewidth]{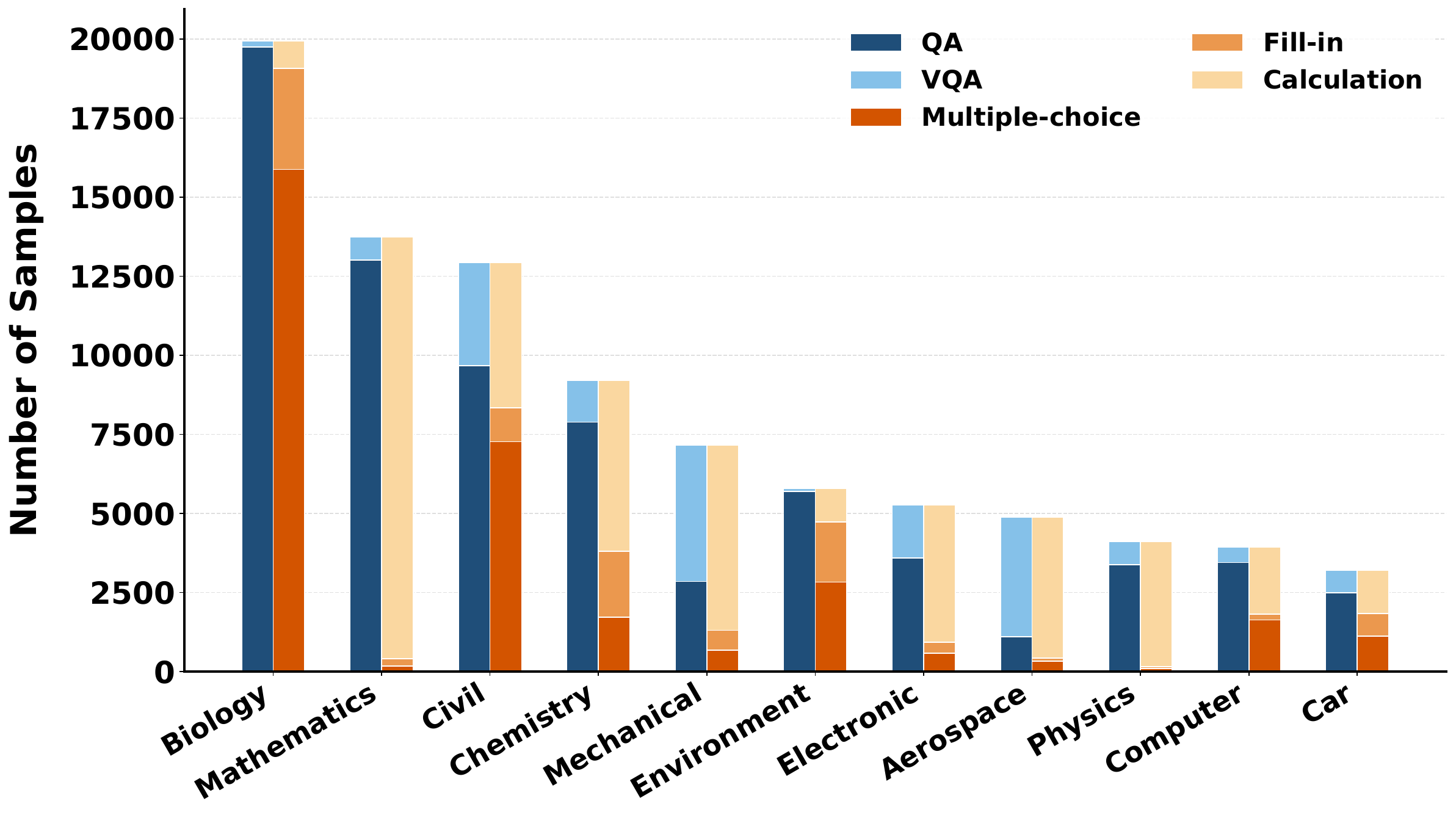}
\caption{Question types and QA/VQA ratio for different subjects in our FlipVQA dataset.}
\label{fig:overall}
\Description{}
\end{figure}

\subsection{Dataset Statistics and Comparisons}
\paragraph{\textbf{Scale and Diversity}} FlipVQA comprises 83,834 questions, including 13,842 VQA pairs and 69,992 QA pairs, which ensures sufficient coverage for fine-tuning large multimodal models. The dataset spans 11 disciplines, fundamental sciences (e.g., Mathematics, Physics) to applied engineering (e.g., Aerospace, Automotive).

\paragraph{\textbf{Cross-Disciplinary Composition}} We maintain multimodal coverage across all subjects, though the VQA-to-QA ratio varies by domain. Fig.~\ref{fig:overall} illustrates Mechanical Engineering and Aerospace exhibit a higher density of visual questions, whereas Biology remains predominantly textual. In terms of task types, FlipVQA is dominated by Calculation (52.5\%), followed by Multiple-choice (35.8\%) and Fill-in-the-blank (11.7\%). The distribution of these tasks is highly subject-dependent; for instance, Calculation is prevalent in formal sciences, while Multiple-choice is more frequent in Biology and Civil Engineering, reflecting the distinct reasoning paradigms of each field.

\section{Experiments and Analysis}

\subsection{Experiment Settings}

\paragraph{\textbf{Models.}}
Our pipeline leverages a suite of specialized models for data processing and synthesis. For structural parsing, we employ MinerU 2.5~\citep{niu2025mineru2} to handle layout recognition and reading order. Gemini-2.5-pro~\citep{comanici2025gemini} serves as the core LLM for logical QA pairing and image insertion. For reasoning-oriented CoT synthesis, we utilize Qwen3-235B-A22B-Thinking (for text-only QA) and Qwen3-VL-235B-A22B-Thinking (for VQA) to generate Chain-of-Thought (CoT) trajectories. We implement a rejection sampling protocol with up to $M=5$ retries, retaining only those trajectories whose final predictions match the ground-truth answers to ensure data fidelity.

\paragraph{\textbf{Benchmarks.}}
We evaluate our method on two fronts: (1) \textbf{Extraction quality}, we select three representative documents with distinct layouts: a solution manual with immediate answers, an abstract algebra textbook with decoupled answers, and a multi-column Chinese workbook with appendix-based answers\footnote{Representative samples and visual layout demonstrations for each document are provided in Appendix~\ref{app:case-study}.}. (2) \textbf{Downstream performance}, we evaluate on nine QA benchmarks and 6 VQA benchmarks, including general reasoning datasets 
(MMLU\citep{hendryckstest2021mmlu},
MMLU-Pro\citep{wang2024mmlu-pro},
BBH\citep{suzgun2022challenging},
SuperGPQA\citep{pteam2025supergpqa},
MMMU\citep{yue2024mmmu},
OlympicArena\citep{huang2024olympicarena})
and domain-specific reasoning benchmarks 
(GSM8K\citep{cobbe2021gsm8k},
MATH500\citep{hendrycks2021math}, 
AIME2024,
AIME2025,
UGPhysics\citep{xu2025ugphysics},
MathVerse\citep{zhang2024mathverse},
MathVista\citep{lumathvista},
OlympiadBench\citep{he2024olympiadbench},
MatSciBench\citep{zhang2025matscibench}
).
Detailed descriptions are provided in Appendix~\ref{app:benchmarks}.

\paragraph{\textbf{Evaluation Protocols.}}
\textbf{Extraction quality} is manually assessed across two modalities: (1) Textual Alignment, verifying the completeness and logical ordering of QA pairs; and (2) Visual Integrity, ensuring correct image placement. We focus on high-level structural errors rather than minor OCR noise, as MinerU 2.5 maintains a high textual F1 score above 0.94. Two human annotators were instructed to label the correctness of the extracted VQA pairs independently, and the inconsistent ones were verified by a third person. Detailed guidelines for annotators are provided in Appendix~\ref{app:annotations}. 

\textbf{Downstream performance} evaluations employ greedy decoding (Temperature=0). Due to response complexity, we adopt an LLM-as-judge protocol using GPT-5-mini~\citep{gpt-5}, following the HLE~\citep{HLE} setup. For the low-sample AIME benchmarks, we report the average of 32 rollouts per question to ensure statistical stability. Prompt details are provided in Appendix~\ref{app:gen_prompt}.

\paragraph{\textbf{Training Protocol.}}
We perform supervised fine-tuning (SFT) on Qwen3-8B-Base (for QA) and Qwen3-VL-8B-Instruct~\citep{bai2025qwen3} (for VQA). For multimodal tuning, we freeze the vision encoder and update only the projector and language backbone. All models are trained using the LLaMA-Factory~\citep{zheng2024llamafactory} framework with DeepSpeed ZeRO-3~\citep{rajbhandari2020zero} acceleration. Hyperparameters are detailed in \cref{app:train_hyperparams}.

\begin{table*}[t]
  \centering
  \caption{Main results on reasoning benchmarks with average scores. \textbf{Top}: Results on textual benchmarks using Qwen3-8B-Base as the base model. \textbf{Bottom}: Results on multimodal benchmarks using Qwen3-VL-8B-Instruct as the base model. Bold indicates the best performance.}
  \label{tab:main_results}
  \renewcommand{\arraystretch}{1.2}
  
  \resizebox{\textwidth}{!}{
    \begin{tabular}{lccccccccccc}
    \toprule
    \multicolumn{12}{l}{\textbf{Textual Reasoning Benchmarks (QA)}} \\
    \midrule
    \textbf{} & \multicolumn{4}{c}{General} & \multicolumn{5}{c}{Math} & Physics & \multirow{2}{*}{\textbf{Average Acc.}} \\
     \cmidrule(lr){2-5} \cmidrule(lr){6-10} \cmidrule(lr){11-11}
    \textbf{Model} & MMLU & MMLU-Pro & BBH & SuperGPQA & GSM8K & MATH500 & AIME24 & AIME25 & AIME26 & UGPhys  \\
    \midrule
    Qwen3-8B-Base & 65.26 & 47.16 & 68.55 & 24.21 & 77.63 & 67.80 & 14.27 & 14.69 & 12.08 & 22.61 & 41.41 \\
    Qwen3-8B (Non-thinking) & 78.78 & 65.39 & 80.99 & 35.90 & 93.40 & 84.80 & 25.31 & \textbf{21.25} & 16.46 & 36.18 & 53.85 \\
    Ours (Qwen3-8B-Base-sft) & \textbf{81.12} & \textbf{67.45} & \textbf{81.37} & \textbf{37.88} & \textbf{95.00} & \textbf{87.40} & \textbf{25.73} & 20.31 & \textbf{18.33} & \textbf{39.95} & \textbf{55.45} \\
    \bottomrule
  \end{tabular}
  }

  \vspace{0.8em} 

  \resizebox{\textwidth}{!}{
  \begin{tabular}{lcccccccc}
    \toprule
    \multicolumn{9}{l}{\textbf{Multimodal Reasoning Benchmarks (VQA)}} \\
  \midrule
  \textbf{} & \multicolumn{3}{c}{General} & Math+Physics & \multicolumn{2}{c}{Math} & Chemistry & \multirow{2}{*}{\textbf{Average Acc.}} \\
   \cmidrule(lr){2-4} \cmidrule(lr){5-5} \cmidrule(lr){6-7} \cmidrule(lr){8-8}
  \textbf{Model} & FlipVQA-test & MMMU & OlympicArena & OlympiadBench & MathVerse & MathVista & MatSciBench  \\
  \midrule
  Qwen3-VL-8B-Instruct & 24.45 & 65.78 & 42.54 & 57.35 & 66.73 & \textbf{70.30} & 29.79 & 50.99 \\
  Ours & \textbf{32.97} & \textbf{66.22} & \textbf{47.39} & \textbf{64.14} & \textbf{71.37} & 70.10 & \textbf{35.62} & \textbf{55.40} \\
  \bottomrule
  \end{tabular}
  }
\end{table*}
\subsection{Performance of FlipVQA-Miner}
\subsubsection{Quantitative Performance}

Table~\ref{tab:extraction-results} presents the quantitative results for both text and vision modalities on the documents. The pipeline achieves text F1 scores above 0.98 across all documents.  Image placement precision is consistently 1.0, with F1 scores above 0.96. These results indicate that our pipeline achieves consistently high extraction accuracy for interleaved sequences, long-distance QA pairs, multi-column layouts, and multilingual content, demonstrating its robustness and potential for broader applications in automated VQA pair extraction. 

To further demonstrate the qualitative results in resolving complex structural ambiguities, we provide a series of representative case studies in Appendix \ref{app:case-study}.

\begin{figure}
\centering
\includegraphics[width=0.95\linewidth]{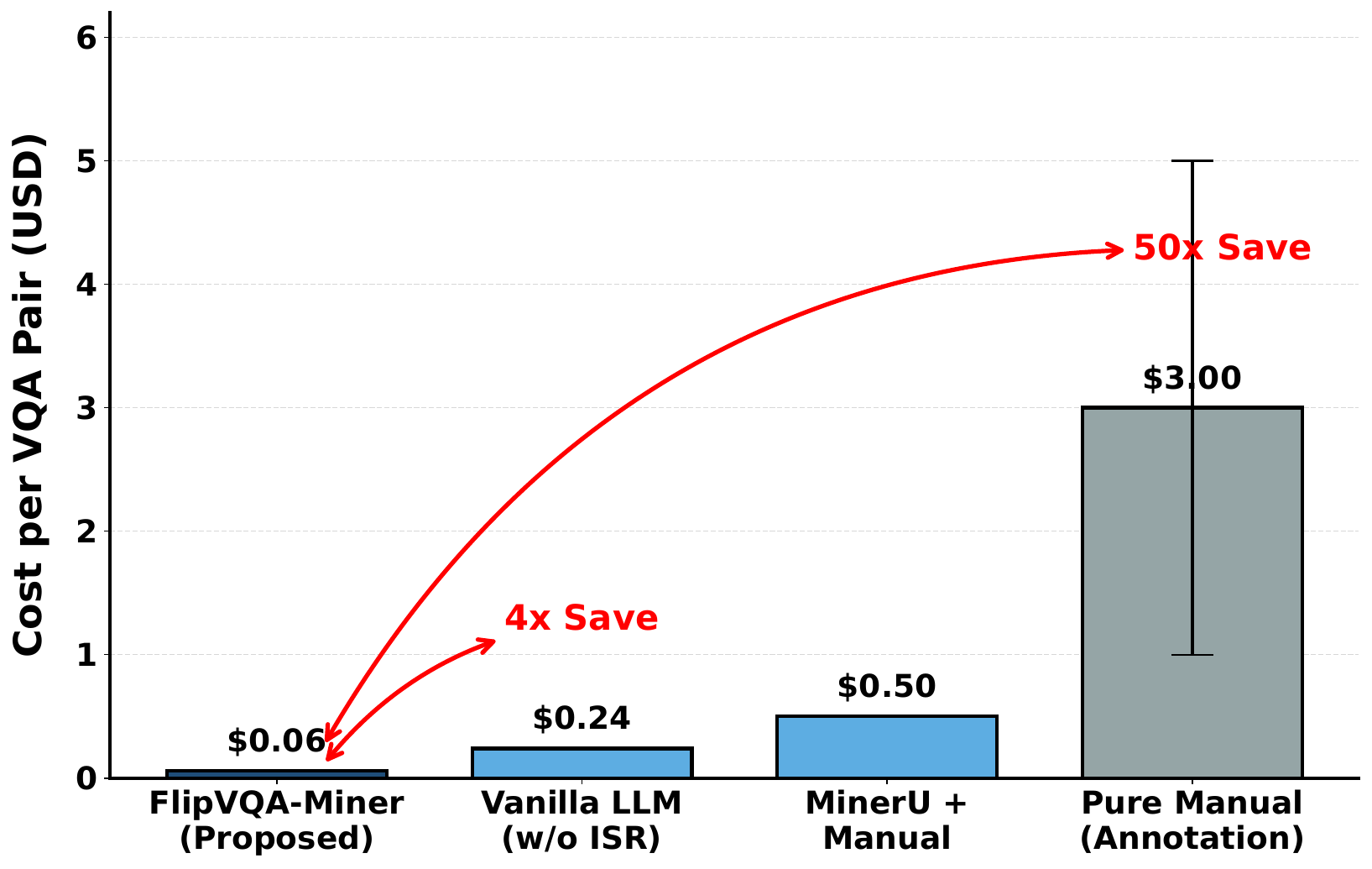}
\caption{Cost-efficiency analysis of our VQA data extraction paradigms.}
        \label{fig:cost}
\Description{}
\end{figure}

\subsubsection{Cost-Efficiency}


Beyond high extraction accuracy, another critical advantage of FlipVQA-Miner lies in its operational economy.
Figure~\ref{fig:cost} illustrates a cost-efficiency analysis of FlipVQA-Miner against alternative VQA data synthesis paradigms. By leveraging the Identifier-guided Semantic Reconstruction (ISR) strategy (\Cref{sec:id_reconstruct}), our pipeline significantly mitigates computational and operational overhead. Specifically, transitioning from holistic LLM-based VQA generation to ISR-driven synthesis reduces the unit cost from \$0.24 to \$0.06, a fourfold ($4\times$) improvement in efficiency over vanilla LLM generation. Furthermore, FlipVQA-Miner proves to be approximately $8\times$ more cost-effective than MinerU-assisted human workflows which cost approximately \$0.50 each pair. Compared to traditional manual annotation, which cost \$1.00 to \$5.00 each pair, FlipVQA-Miner yields an average $50\times$ reduction in expenditure, highlighting its economic scalability for the curation of large-scale multimodal reasoning benchmarks.

\subsection{Performance of Various Models Trained on FlipVQA-83K Dataset}

The main results are summarized in Table~\ref{tab:main_results}. Overall, our method consistently outperforms the baselines across both textual and multimodal reasoning benchmarks.

On textual benchmarks, our fine-tuned model improves performance on 9 out of 10 datasets compared with the Qwen3-8B Non-thinking mode. In particular, notable gains are observed on MMLU, Math500, and UGPhysics, indicating improvements in both general reasoning ability and domain knowledge in scientific subjects.

On multimodal benchmarks, our model achieves improvements on 5 out of 7 datasets compared with the Qwen3-VL-8B baseline, showing enhanced visual reasoning performance across disciplines such as mathematics, physics, and chemistry. On MMMU, the model exhibits a modest improvement, indicating that the specialization introduced by fine-tuning does not compromise its general-purpose multimodal capability.

A slight performance decrease is observed on MathVista, likely due to distributional differences between the training and evaluation data. Specifically, FlipVQA-83K mainly consists of problems extracted from college-level textbooks, whereas MathVista focuses more on logic puzzles, functional plots, and figures commonly found in academic publications.

\vspace{-1em}
\subsection{Ablation Study: Impact of Data Curation}
To quantify the impact of our \textbf{Multi-stage Data Curation Pipeline}, we compare the final FlipVQA model against a baseline trained on raw extractions from the FlipVQA-Miner. As shown in Table~\ref{tab:ablation}, our data curation significantly outperforms the raw extraction baseline. While textbook corpora provide high fidelity problem and answer pairs, enhancing latent reasoning requires meticulously curated long form CoT trajectories. Without multi-stage data curation, training on raw data lacking structured reasoning fails to surpass or may even degrade the inherent reasoning priors of Qwen3-VL-8B-Instruct. Thus, rigorous data curation and CoT synthesis are indispensable for distilling textbook knowledge into executable multimodal intelligence.


\begin{table}[h]
\centering
\small
\caption{Ablation study of the FlipVQA curation pipeline. Phase 1 refers to FlipVQA-Miner and Phase 2 represents the Multi-stage Data Curation Pipeline.}

\label{tab:ablation_qa}
\resizebox{0.48\textwidth}{!}{
\begin{tabular}{l|cc|cc|c}
\toprule
\multicolumn{6}{l}{\textbf{Textual Reasoning Benchmarks (QA)}} \\
\midrule
\textbf{Model} & \textbf{Phase 1} & \textbf{Phase 2} & \textbf{MMLU-Pro} & \textbf{SuperGPQA} & \textbf{Avg.} \\ \midrule
Qwen3-8B-Base & \(\times\) & \(\times\) & 47.16 & 24.21 & 35.69 \\
Qwen3-8B(Non-thinking) & \(\times\) & \(\times\) & 65.39 & 35.90 & 50.65 \\
Raw extraction only & \(\checkmark\) & \(\times\) & 32.11 & 15.53 & 23.82 \\
\rowcolor[gray]{0.9} \textbf{FlipVQA-83K(QA)} & \(\checkmark\) & \(\checkmark\) & \textbf{67.11} & \textbf{37.39} & \textbf{52.25} \\ \bottomrule
\end{tabular}
}

\vspace{1em}

\label{tab:ablation}
\resizebox{0.48\textwidth}{!}{
\begin{tabular}{l|cc|cc|c}
\toprule
\multicolumn{6}{l}{\textbf{Multimodal Reasoning Benchmarks (VQA)}} \\
\midrule
\textbf{Model} & \textbf{Phase 1} & \textbf{Phase 2} & \textbf{MMMU} & \textbf{OlympicArena} & \textbf{Avg.} \\ \midrule
Qwen3-VL-8B-Instruct & \(\times\) & \(\times\) & 65.78 & 42.54 & 54.16 \\
Raw extraction only & \(\checkmark\) & \(\times\) & 53.67 & 30.97 & 42.32 \\
\rowcolor[gray]{0.9} \textbf{FlipVQA-83K(VQA)} & \(\checkmark\) & \(\checkmark\) & \textbf{66.22} & \textbf{47.39} & \textbf{56.81} \\ \bottomrule
\end{tabular}
}


\end{table}

\vspace{-2em}

\section{Conclusion}
We have presented \textbf{FlipVQA-Miner}, an automated pipeline that synthesizes high-fidelity, reasoning-oriented VQA pairs from multi-disciplinary textbooks. By effectively resolving long-range logical dependencies and cross-page discontinuities inherent in pedagogical documents, our framework transforms raw content into grounded, hallucination-free supervision signals. We have demonstrated that FlipVQA-Miner achieves a \textbf{50$\times$} cost saving compared to manual annotation while maintaining exceptional structural fidelity ($F_1 > 0.96$). Our multi-stage data curation pipeline is critical for transforming raw extractions into grounded supervision signals through curated long form Chain of Thought trajectories. Utilizing this pipeline, we released the \textbf{FlipVQA-83K}, comprising 83k instances spanning 11 academic disciplines. Models fine-tuned on FlipVQA-83K show significantly improved reasoning robustness, surpassing stronger instruction-tuned baselines across both textual and multimodal benchmarks. 

\begin{acks}
This work is supported by National Natural Science Foundation of China (92470121, 62402016), Fundamental and Interdisciplinary Disciplines Breakthrough Plan of the Ministry of Education of China (JYB2025XDXM113), National Key R\&D Program of China (2024YFA1014003), Zhongguancun Academy (C20250204, C20250602),  Beijing Major Science and Technology Project (Z251100008125043, Z251100008425023), and High-performance Computing Platform of Peking University.
\end{acks}

\bibliographystyle{ACM-Reference-Format}
\bibliography{custom}

@article{guo2025deepseek,
  title={Deepseek-r1: Incentivizing reasoning capability in llms via reinforcement learning},
  author={Guo, Daya and Yang, Dejian and Zhang, Haowei and Song, Junxiao and Wang, Peiyi and Zhu, Qihao and Xu, Runxin and Zhang, Ruoyu and Ma, Shirong and Bi, Xiao and others},
  journal={arXiv preprint arXiv:2501.12948},
  year={2025}
}

@inproceedings{wang2023self,
  title={Self-instruct: Aligning language models with self-generated instructions},
  author={Wang, Yizhong and Kordi, Yeganeh and Mishra, Swaroop and Liu, Alisa and Smith, Noah A and Khashabi, Daniel and Hajishirzi, Hannaneh},
  booktitle={Proceedings of the 61st annual meeting of the association for computational linguistics (volume 1: long papers)},
  pages={13484--13508},
  year={2023}
}

@inproceedings{tan2024large,
  title={Large language models for data annotation and synthesis: A survey},
  author={Tan, Zhen and Li, Dawei and Wang, Song and Beigi, Alimohammad and Jiang, Bohan and Bhattacharjee, Amrita and Karami, Mansooreh and Li, Jundong and Cheng, Lu and Liu, Huan},
  booktitle={Proceedings of the 2024 Conference on Empirical Methods in Natural Language Processing},
  pages={930--957},
  year={2024}
}

@inproceedings{huang2022layoutlmv3,
  title={Layoutlmv3: Pre-training for document ai with unified text and image masking},
  author={Huang, Yupan and Lv, Tengchao and Cui, Lei and Lu, Yutong and Wei, Furu},
  booktitle={Proceedings of the 30th ACM international conference on multimedia},
  pages={4083--4091},
  year={2022}
}

@article{niu2025mineru2,
  title={Mineru2. 5: A decoupled vision-language model for efficient high-resolution document parsing},
  author={Niu, Junbo and Liu, Zheng and Gu, Zhuangcheng and Wang, Bin and Ouyang, Linke and Zhao, Zhiyuan and Chu, Tao and He, Tianyao and Wu, Fan and Zhang, Qintong and others},
  journal={arXiv preprint arXiv:2509.22186},
  year={2025}
}

@article{zhao2024doclayout,
  title={Doclayout-yolo: Enhancing document layout analysis through diverse synthetic data and global-to-local adaptive perception},
  author={Zhao, Zhiyuan and Kang, Hengrui and Wang, Bin and He, Conghui},
  journal={arXiv preprint arXiv:2410.12628},
  year={2024}
}

@inproceedings{lee2023pix2struct,
  title={Pix2struct: Screenshot parsing as pretraining for visual language understanding},
  author={Lee, Kenton and Joshi, Mandar and Turc, Iulia Raluca and Hu, Hexiang and Liu, Fangyu and Eisenschlos, Julian Martin and Khandelwal, Urvashi and Shaw, Peter and Chang, Ming-Wei and Toutanova, Kristina},
  booktitle={International Conference on Machine Learning},
  pages={18893--18912},
  year={2023},
  organization={PMLR}
}

@article{comanici2025gemini,
  title={Gemini 2.5: Pushing the frontier with advanced reasoning, multimodality, long context, and next generation agentic capabilities},
  author={Comanici, Gheorghe and Bieber, Eric and Schaekermann, Mike and Pasupat, Ice and Sachdeva, Noveen and Dhillon, Inderjit and Blistein, Marcel and Ram, Ori and Zhang, Dan and Rosen, Evan and others},
  journal={arXiv preprint arXiv:2507.06261},
  year={2025}
}

@inproceedings{yue2024mmmu,
  title={Mmmu: A massive multi-discipline multimodal understanding and reasoning benchmark for expert agi},
  author={Yue, Xiang and Ni, Yuansheng and Zhang, Kai and Zheng, Tianyu and Liu, Ruoqi and Zhang, Ge and Stevens, Samuel and Jiang, Dongfu and Ren, Weiming and Sun, Yuxuan and others},
  booktitle={Proceedings of the IEEE/CVF conference on computer vision and pattern recognition},
  pages={9556--9567},
  year={2024}
}

@article{huang2024olympicarena,
  title={Olympicarena: Benchmarking multi-discipline cognitive reasoning for superintelligent ai},
  author={Huang, Zhen and Wang, Zengzhi and Xia, Shijie and Li, Xuefeng and Zou, Haoyang and Xu, Ruijie and Fan, Run-Ze and Ye, Lyumanshan and Chern, Ethan and Ye, Yixin and others},
  journal={Advances in Neural Information Processing Systems},
  volume={37},
  pages={19209--19253},
  year={2024}
}

@inproceedings{zhang2024mathverse,
  title={Mathverse: Does your multi-modal llm truly see the diagrams in visual math problems?},
  author={Zhang, Renrui and Jiang, Dongzhi and Zhang, Yichi and Lin, Haokun and Guo, Ziyu and Qiu, Pengshuo and Zhou, Aojun and Lu, Pan and Chang, Kai-Wei and Qiao, Yu and others},
  booktitle={European Conference on Computer Vision},
  pages={169--186},
  year={2024},
  organization={Springer}
}

@article{lumathvista,
  title={Mathvista: Evaluating mathematical reasoning of foundation models in visual contexts},
  author={Lu, Pan and Bansal, Hritik and Xia, Tony and Liu, Jiacheng and Li, Chunyuan and Hajishirzi, Hannaneh and Cheng, Hao and Chang, Kai-Wei and Galley, Michel and Gao, Jianfeng},
  journal={arXiv preprint arXiv:2310.02255},
  year={2023}
}

@inproceedings{he2024olympiadbench,
  title={Olympiadbench: A challenging benchmark for promoting agi with olympiad-level bilingual multimodal scientific problems},
  author={He, Chaoqun and Luo, Renjie and Bai, Yuzhuo and Hu, Shengding and Thai, Zhen and Shen, Junhao and Hu, Jinyi and Han, Xu and Huang, Yujie and Zhang, Yuxiang and others},
  booktitle={Proceedings of the 62nd Annual Meeting of the Association for Computational Linguistics (Volume 1: Long Papers)},
  pages={3828--3850},
  year={2024}
}

@article{zhang2025matscibench,
  title={MatSciBench: Benchmarking the Reasoning Ability of Large Language Models in Materials Science},
  author={Zhang, Junkai and Gan, Jingru and Wang, Xiaoxuan and Jia, Zian and Gu, Changquan and Chen, Jianpeng and Zhu, Yanqiao and Ma, Mingyu Derek and Zhou, Dawei and Li, Ling and others},
  journal={arXiv preprint arXiv:2510.12171},
  year={2025}
}

@article{bai2025qwen3,
  title={Qwen3-vl technical report},
  author={Bai, Shuai and Cai, Yuxuan and Chen, Ruizhe and Chen, Keqin and Chen, Xionghui and Cheng, Zesen and Deng, Lianghao and Ding, Wei and Gao, Chang and Ge, Chunjiang and others},
  journal={arXiv preprint arXiv:2511.21631},
  year={2025}
}

@inproceedings{zheng2024llamafactory,
  title={Llamafactory: Unified efficient fine-tuning of 100+ language models},
  author={Zheng, Yaowei and Zhang, Richong and Zhang, Junhao and Ye, Yanhan and Luo, Zheyan},
  booktitle={Proceedings of the 62nd annual meeting of the association for computational linguistics (volume 3: system demonstrations)},
  pages={400--410},
  year={2024}
}

@inproceedings{rajbhandari2020zero,
  title={Zero: Memory optimizations toward training trillion parameter models},
  author={Rajbhandari, Samyam and Rasley, Jeff and Ruwase, Olatunji and He, Yuxiong},
  booktitle={SC20: international conference for high performance computing, networking, storage and analysis},
  pages={1--16},
  year={2020},
  organization={IEEE}
}

@article{HLE,
  title={A benchmark of expert-level academic questions to assess AI capabilities},
  author={Center for AI Safety Phan Long agibenchmark@ safe. ai 1 Gatti Alice 1 Li Nathaniel 1 Khoja Adam 1 Kim Ryan 1 Ren Richard 1 Hausenloy Jason 1 Zhang Oliver 1 Mazeika Mantas 1 Hendrycks Dan dan@ safe. ai 1},
  journal={Nature},
  volume={649},
  number={8099},
  pages={1139--1146},
  year={2026},
  publisher={Nature Publishing Group UK London}
}

@article{gpt-5,
  title={Openai gpt-5 system card},
  author={Singh, Aaditya and Fry, Adam and Perelman, Adam and Tart, Adam and Ganesh, Adi and El-Kishky, Ahmed and McLaughlin, Aidan and Low, Aiden and Ostrow, AJ and Ananthram, Akhila and others},
  journal={arXiv preprint arXiv:2601.03267},
  year={2025}
}

@article{hendryckstest2021mmlu,
  title={Measuring massive multitask language understanding},
  author={Hendrycks, Dan and Burns, Collin and Basart, Steven and Zou, Andy and Mazeika, Mantas and Song, Dawn and Steinhardt, Jacob},
  journal={arXiv preprint arXiv:2009.03300},
  year={2020}
}

@article{wang2024mmlu-pro,
  title={Mmlu-pro: A more robust and challenging multi-task language understanding benchmark},
  author={Wang, Yubo and Ma, Xueguang and Zhang, Ge and Ni, Yuansheng and Chandra, Abhranil and Guo, Shiguang and Ren, Weiming and Arulraj, Aaran and He, Xuan and Jiang, Ziyan and others},
  journal={Advances in Neural Information Processing Systems},
  volume={37},
  pages={95266--95290},
  year={2024}
}

@inproceedings{suzgun2022challenging,
  title={Challenging big-bench tasks and whether chain-of-thought can solve them},
  author={Suzgun, Mirac and Scales, Nathan and Sch{\"a}rli, Nathanael and Gehrmann, Sebastian and Tay, Yi and Chung, Hyung Won and Chowdhery, Aakanksha and Le, Quoc and Chi, Ed and Zhou, Denny and others},
  booktitle={Findings of the Association for Computational Linguistics: ACL 2023},
  pages={13003--13051},
  year={2023}
}

@article{pteam2025supergpqa,
  title={Supergpqa: Scaling llm evaluation across 285 graduate disciplines},
  author={Du, Xinrun and Yao, Yifan and Ma, Kaijing and Wang, Bingli and Zheng, Tianyu and Zhu, King and Liu, Minghao and Liang, Yiming and Jin, Xiaolong and Wei, Zhenlin and others},
  journal={arXiv preprint arXiv:2502.14739},
  year={2025}
}

@article{cobbe2021gsm8k,
  title={Training verifiers to solve math word problems},
  author={Cobbe, Karl and Kosaraju, Vineet and Bavarian, Mohammad and Chen, Mark and Jun, Heewoo and Kaiser, Lukasz and Plappert, Matthias and Tworek, Jerry and Hilton, Jacob and Nakano, Reiichiro and others},
  journal={arXiv preprint arXiv:2110.14168},
  year={2021}
}

@article{hendrycks2021math,
  title={Measuring mathematical problem solving with the math dataset},
  author={Hendrycks, Dan and Burns, Collin and Kadavath, Saurav and Arora, Akul and Basart, Steven and Tang, Eric and Song, Dawn and Steinhardt, Jacob},
  journal={arXiv preprint arXiv:2103.03874},
  year={2021}
}

@article{xu2025ugphysics,
  title={Ugphysics: A comprehensive benchmark for undergraduate physics reasoning with large language models},
  author={Xu, Xin and Xu, Qiyun and Xiao, Tong and Chen, Tianhao and Yan, Yuchen and Zhang, Jiaxin and Diao, Shizhe and Yang, Can and Wang, Yang},
  journal={arXiv preprint arXiv:2502.00334},
  year={2025}
}

@inproceedings{smith2007tesseract,
  title={An Overview of the Tesseract OCR Engine},
  author={Smith, Ray},
  booktitle={Ninth International Conference on Document Analysis and Recognition (ICDAR 2007)},
  pages={629--633},
  year={2007}
}

@inproceedings{xu2020layoutlm,
  title={Layoutlm: Pre-training of text and layout for document image understanding},
  author={Xu, Yiheng and Li, Minghao and Cui, Lei and Huang, Shaohan and Wei, Furu and Zhou, Ming},
  booktitle={Proceedings of the 26th ACM SIGKDD international conference on knowledge discovery \& data mining},
  pages={1192--1200},
  year={2020}
}

@inproceedings{kim2022donut,
  title={Ocr-free document understanding transformer},
  author={Kim, Geewook and Hong, Teakgyu and Yim, Moonbin and Nam, JeongYeon and Park, Jinyoung and Yim, Jinyeong and Hwang, Wonseok and Yun, Sangdoo and Han, Dongyoon and Park, Seunghyun},
  booktitle={European Conference on Computer Vision},
  pages={498--517},
  year={2022},
  organization={Springer}
}

@article{blecher2023nougat,
  title={Nougat: Neural optical understanding for academic documents},
  author={Blecher, Lukas and Cucurull, Guillem and Scialom, Thomas and Stojnic, Robert},
  journal={arXiv preprint arXiv:2308.13418},
  year={2023}
}

@inproceedings{kembhavi2017tqa,
  title={Are you smarter than a sixth grader? textbook question answering for multimodal machine comprehension},
  author={Kembhavi, Aniruddha and Seo, Minjoon and Schwenk, Dustin and Choi, Jonghyun and Farhadi, Ali and Hajishirzi, Hannaneh},
  booktitle={Proceedings of the IEEE Conference on Computer Vision and Pattern recognition},
  pages={4999--5007},
  year={2017}
}

@article{lu2022scienceqa,
  title={Learn to explain: Multimodal reasoning via thought chains for science question answering},
  author={Lu, Pan and Mishra, Swaroop and Xia, Tanglin and Qiu, Liang and Chang, Kai-Wei and Zhu, Song-Chun and Tafjord, Oyvind and Clark, Peter and Kalyan, Ashwin},
  journal={Advances in neural information processing systems},
  volume={35},
  pages={2507--2521},
  year={2022}
}

@article{liu2023llava,
  title={Visual instruction tuning},
  author={Liu, Haotian and Li, Chunyuan and Wu, Qingyang and Lee, Yong Jae},
  journal={Advances in neural information processing systems},
  volume={36},
  pages={34892--34916},
  year={2023}
}

@article{li2023m3it,
  title={M3IT: A Large-Scale Dataset towards Multi-Modal Multilingual Instruction Tuning},
  author={Li, Lei and Yin, Yuwei and Li, Shicheng and Chen, Liang and Wang, Peiyi and Ren, Shuhuai and Li, Mukai and Yang, Yazheng and Xu, Jingjing and Sun, Xu and others},
  journal={arXiv preprint arXiv:2306.04387},
  year={2023}
}

\clearpage
\appendix
\onecolumn

\section{Case Studies of VQA Pair Extraction}

\label{app:case-study}

To demonstrate the robustness of our pipeline in resolving diverse structural ambiguities commonly found in educational materials, we present a set of representative examples. Each example includes the original question, the original solution(s), and the corresponding extracted QA or VQA pair rendered in Markdown. These cases illustrate how the system reconstructs raw, unstructured content into a coherent, machine-readable format.

\subsection{Long-distance, cross-document VQA association case}
\begin{figure}[ht]
\centering
\includegraphics[width=0.95\linewidth]{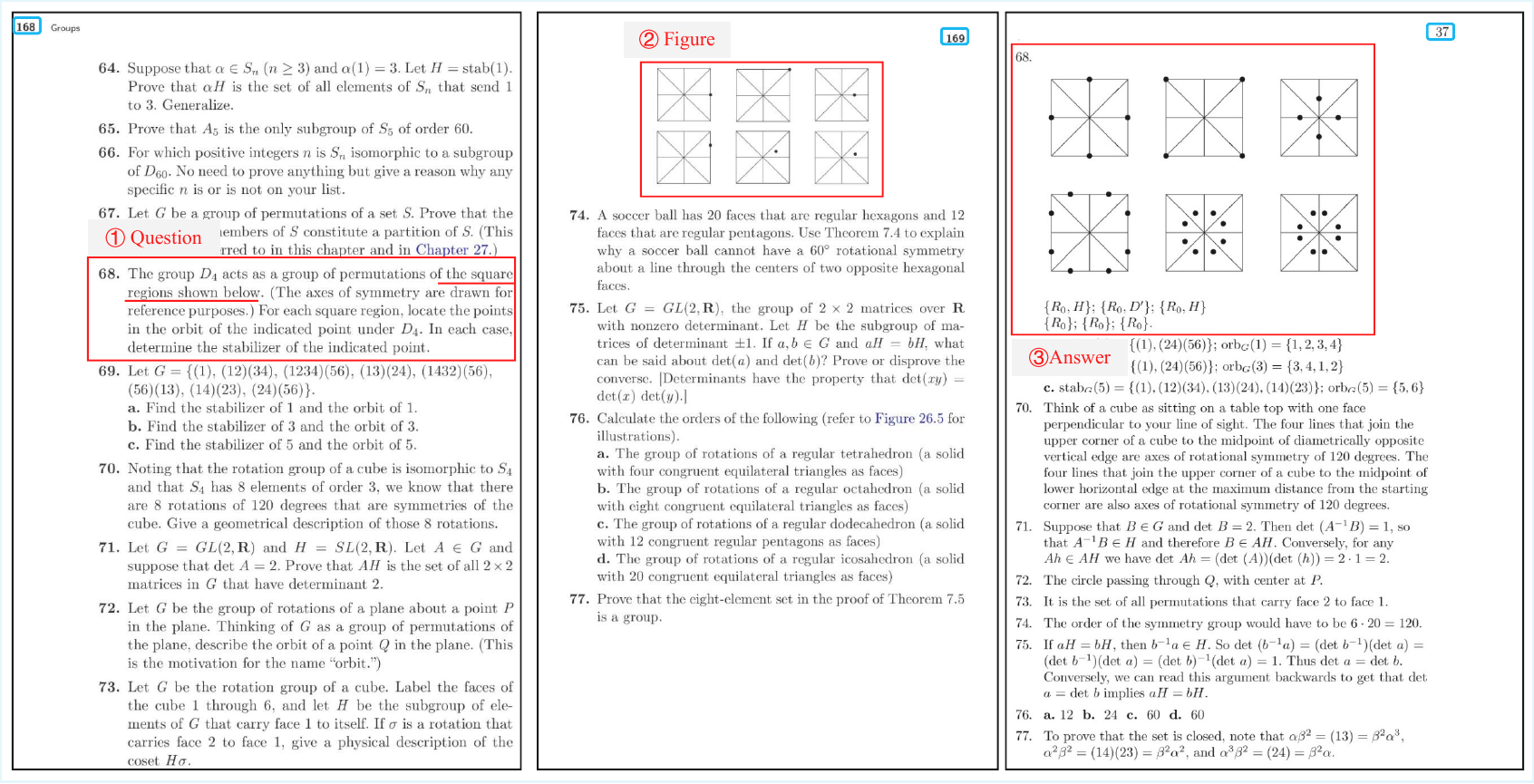}
\caption{Original arrangement in PDF where the question, figure, and answer are separated across pages and documents.}
\label{fig:case1_combined_ori}
\end{figure}

\begin{figure}[ht]
\centering
\includegraphics[width=0.95\linewidth]{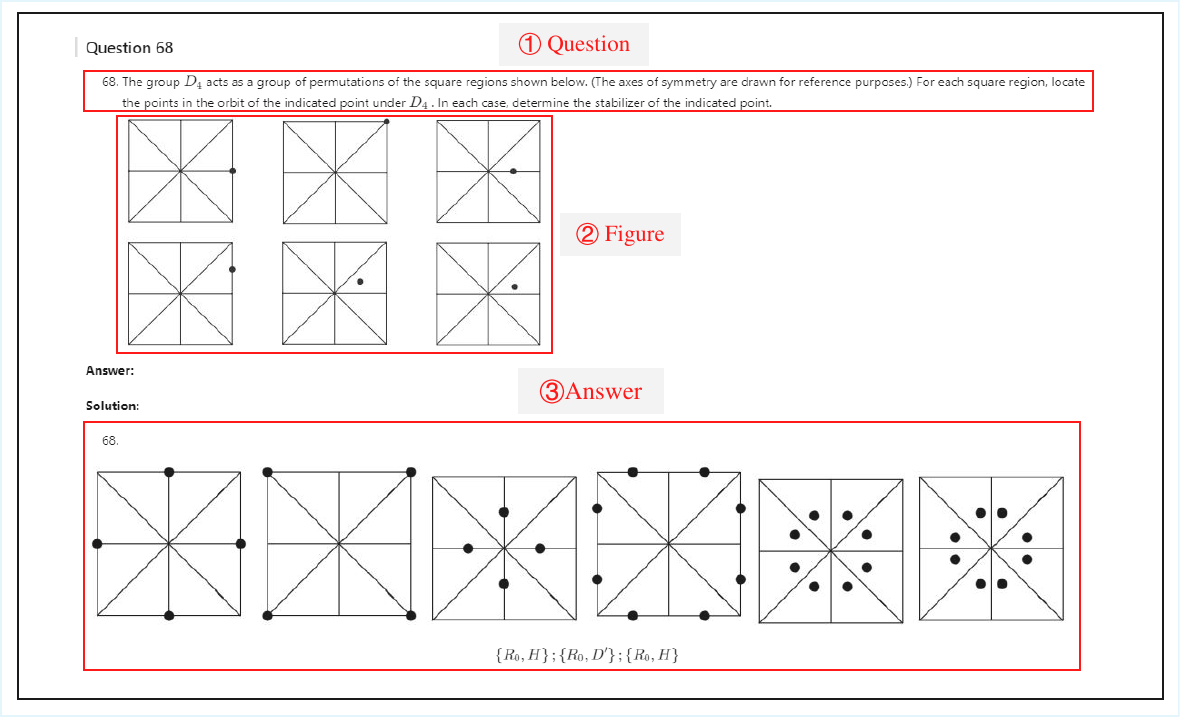}
\caption{Extracted and reconstructed VQA pair rendered in Markdown, showing successful long-distance grouping.}
\label{fig:case1_combined_extract}
\end{figure}

This case demonstrates FlipVQA-Miner’s capability to resolve extreme structural decoupling where components of a single VQA triplet are distributed across distant pages and even different materials. As illustrated in Figure~\ref{fig:case1_combined_ori}, the question (\ding{172}) refers to specific "square regions", yet the associated figure (\ding{173}) is located on a subsequent page and lacks any explicit caption or label. Furthermore, the corresponding answer (\ding{174}) is found in a separate companion volume rather than the primary document. This geographic separation poses a significant challenge for conventional extraction tools. Despite these hurdles, FlipVQA-Miner successfully identifies the latent semantic links, effectively reassembling the fragmented elements into a unified, structured VQA unit. The final Markdown-rendered output, showcasing the system's ability to maintain logical coherence across long-distance associations, is shown in Figure~\ref{fig:case1_combined_extract}.

\clearpage

\subsection{Interleaved QA with multiple solutions case}
\begin{figure}[ht]
    \centering
    \includegraphics[width=0.9\linewidth]{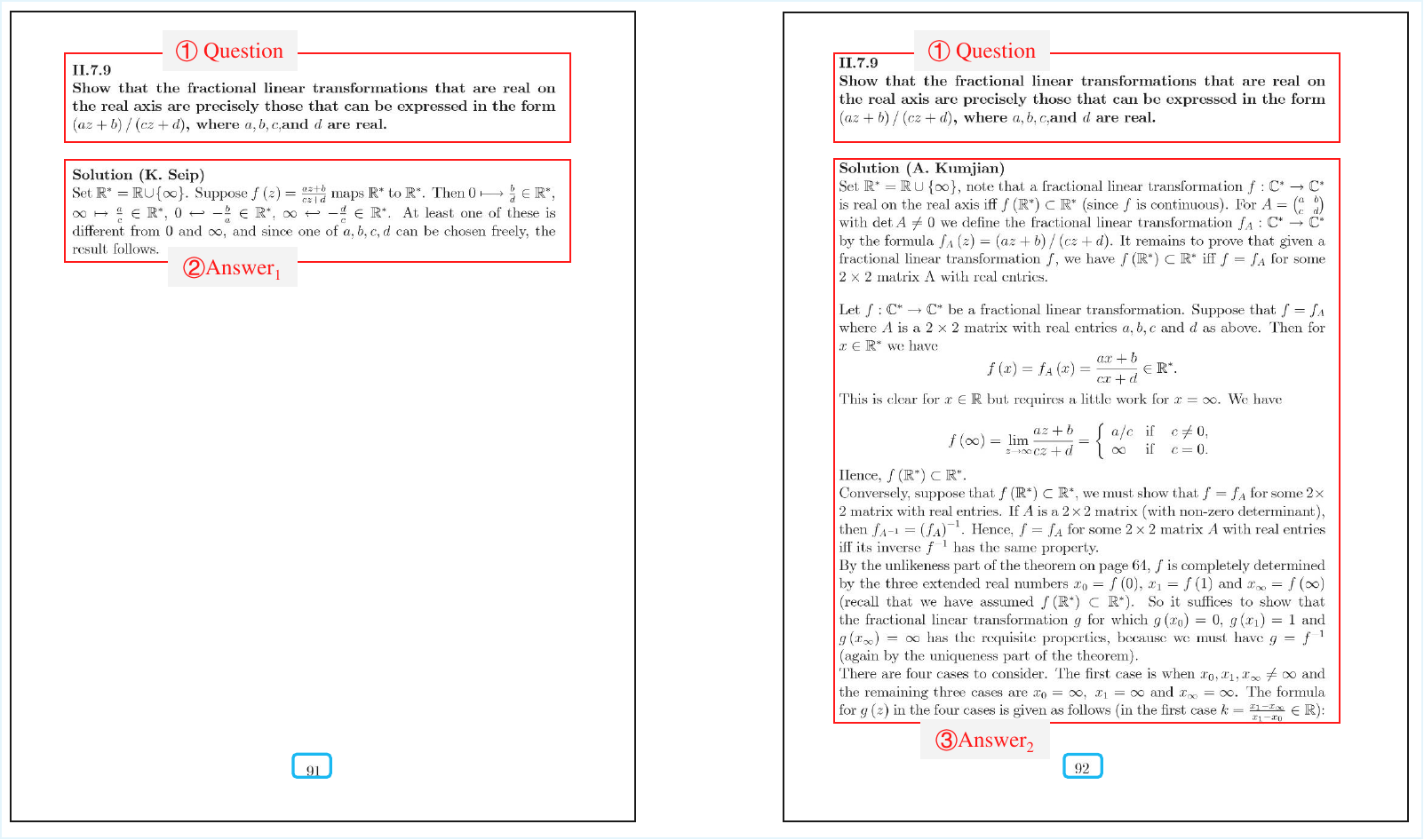}
    \caption{Original question with two interleaved solutions.}
    \label{fig:case2_qa_ori}
\end{figure}

\begin{figure}[ht]
    \centering
    \includegraphics[width=0.9\linewidth]{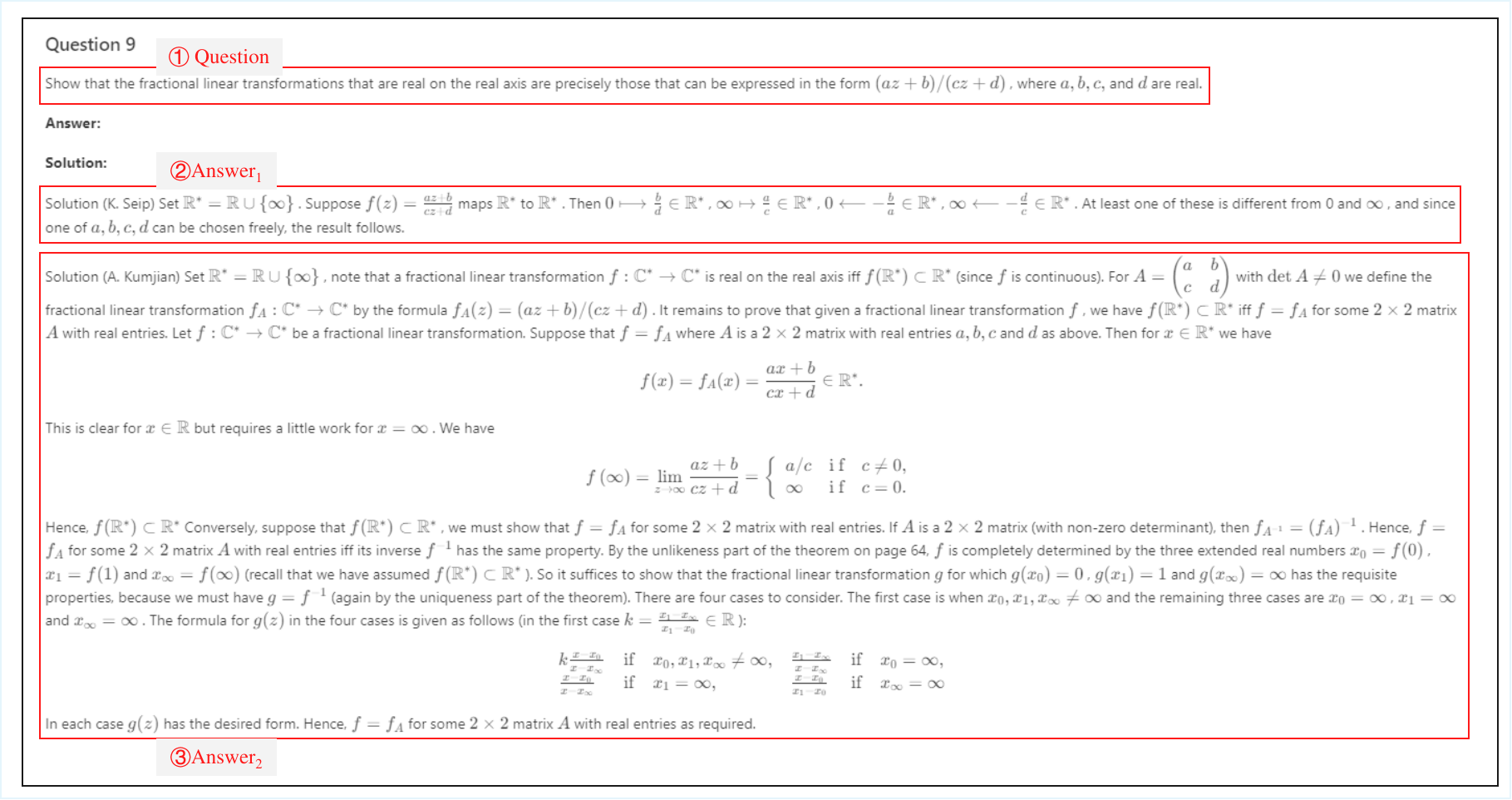}
    \caption{Extracted and reconstructed QA pair rendered in Markdown.}
    \label{fig:case2_qa_extract}
\end{figure}

This case features an interleaved question (Figure~\ref{fig:case2_qa_ori}) accompanied by two distinct solution paths. Despite the intertwined layout, our pipeline faithfully preserves both solutions and correctly associates them with the question. The resulting structured output is shown in Figure~\ref{fig:case2_qa_extract}.

\clearpage

\subsection{Multi-column, cross-page Chinese exercise book case}
\begin{figure}[ht]
    \centering
    \includegraphics[width=\linewidth]{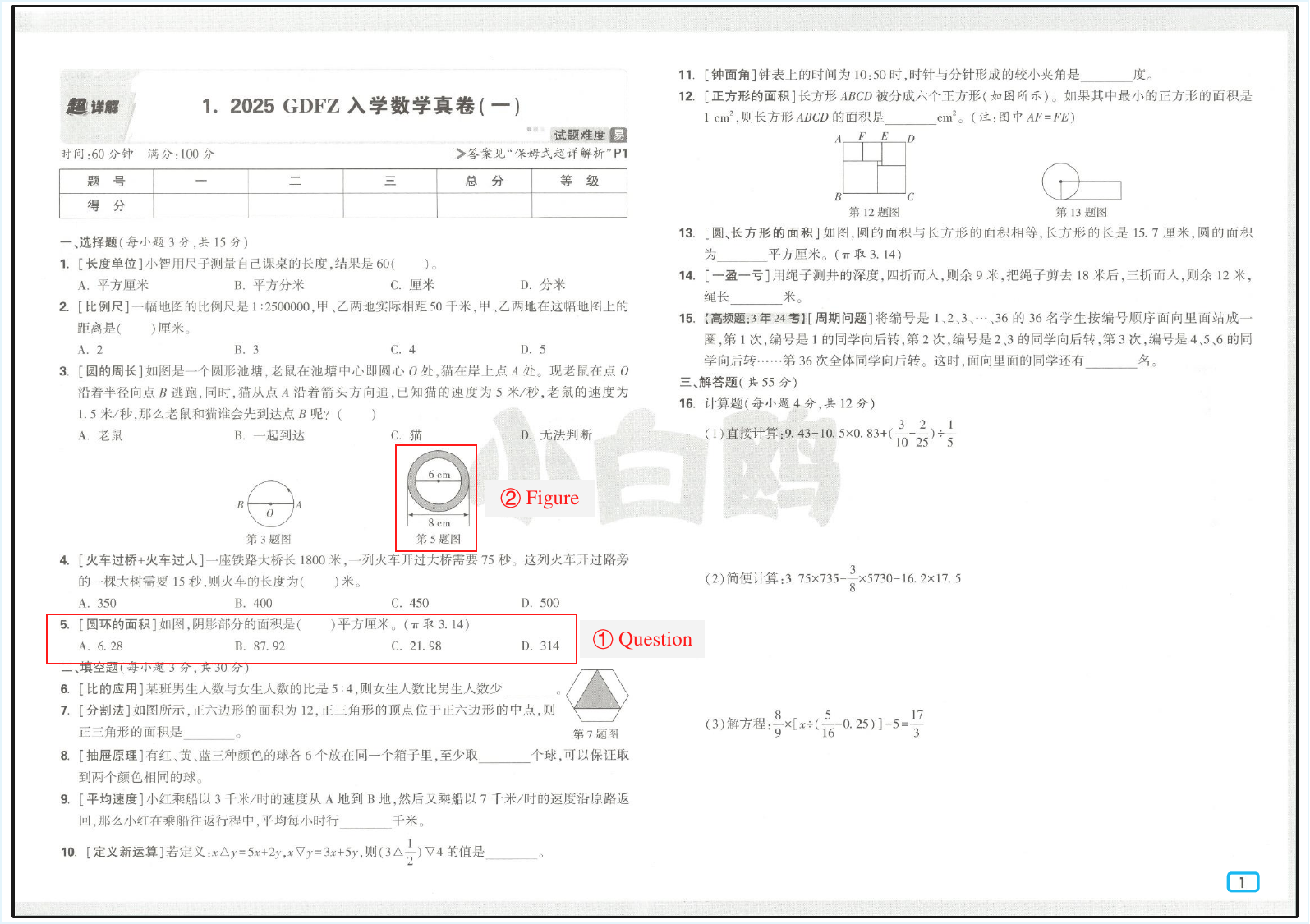}
    \caption{Original question within a multi-column Chinese exercise book layout.}
    \label{fig:case3_q_ori}
\end{figure}

\begin{figure}[ht]
    \centering
    \includegraphics[width=\linewidth]{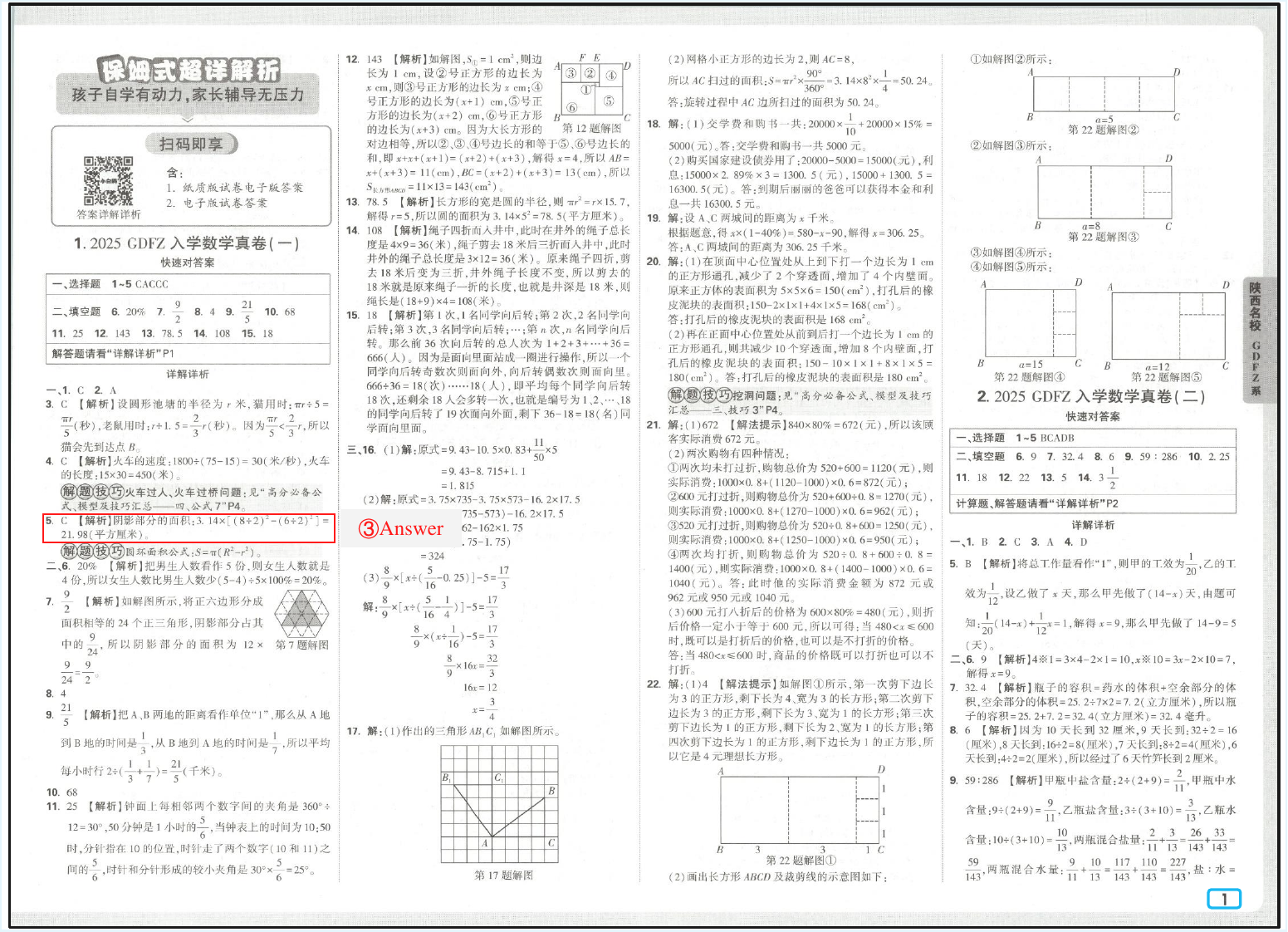}
    \caption{Original answer located in a separate multi-column document.}
    \label{fig:case3_a_ori}
\end{figure}

\begin{figure}[ht]
    \centering
    \includegraphics[width=\linewidth]{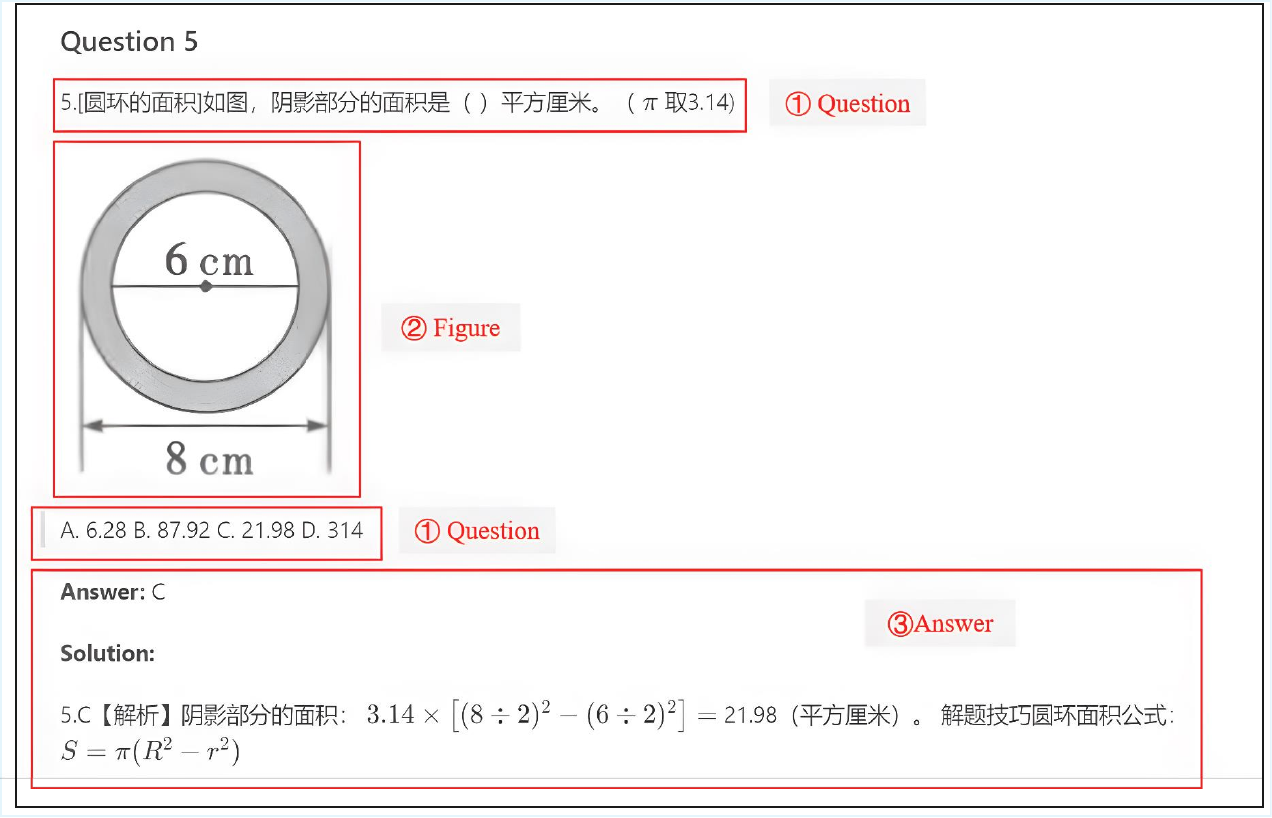}
    \caption{Extracted and reconstructed VQA pair rendered in Markdown.}
    \label{fig:case3_qa_extract}
\end{figure}

This example showcases a particularly challenging scenario drawn from a Chinese exercise book. As shown in Figure~\ref{fig:case3_q_ori}, the question appears within a dense multi-column layout interspersed with unrelated textual elements. Complicating matters further, the corresponding answer (Figure~\ref{fig:case3_a_ori}) is located in a separate PDF file altogether. Despite these structural and formatting hurdles, our pipeline accurately identifies, extracts, and associates the VQA pair. The final Markdown-rendered output (Figure~\ref{fig:case3_qa_extract}) highlights the system’s resilience to multi-column layouts, cross-page and cross-document separation, heterogeneous formatting conventions, and multilingual content.

\clearpage

\section{Prompts Used in Dataset Curation}
\subsection{Identifier-Guided Semantic Reconstruction of VQA Pairs (\Cref{sec:id_reconstruct})}
\label{app:id_reconstruct}
The prompt used for LLM to reconstruct VQA Pairs from the PDF layout with block identifiers is as follows:
\begin{promptbox}[title=VQA Pairs Reconstruction]
\begin{lstlisting}[basicstyle=\ttfamily\scriptsize\linespread{1.0}\selectfont,
                    breaklines=true,
                    breakatwhitespace=false,
                    columns=flexible,
                    xleftmargin=0pt,
                    breakindent=0pt]
You are an expert in {subject}. You are given a json file. Your task is to segment the content, insert images tags, and extract labels:
1. Every json item has an "id" field. Your main task is to output this field.
2. You need to segment the content into multiple `<qa_pair>`...`</qa_pair>` blocks, each containing a question and its corresponding answer with solution.
3. If the problem or answer is not complete, omit them.
4. You need to put the images id into proper positions. You could look at the caption or context to decide where to put the image tags.
5. You will also need to extract the chapter title and each problem's label/number from the text.
6. You only need to output "id" field for chapters, questions and solutions. DO NOT OUTPUT ORIGINAL TEXT. Use ',' to separate different ids.
7. However, use original labels/numbers for labels, and use original numbers for answers. DO NOT output "id" field for labels and answers. You will need to extract them from the text.

Strict extraction rules:
** About questions and answers/solutions **
- Preserve each problem's original label/number, such as "Example 3", "Exercise 1", "11". Do not include the period after the number. Use Arabic numerals only. For example, if the label is "IV", convert it to "4". 
- If the full label is "III.16", keep only "16". If the full label is "5.4", keep only "4".
- If there are multiple sub-questions (such as "(1)", "(a)") under one main question, always put them together in the same `<qa_pair>`...`</qa_pair>` block.
- If a question and its answer/solution are contiguous, wrap them together as a single `<qa_pair>`...`</qa_pair>` block, e.g.:
  `<qa_pair><label>Example 1</label><question>...</question><answer>...</answer><solution>...</solution></qa_pair>`
- If only questions or only answers with solutions appear, wrap each question or answer with solution in a `<qa_pair>`...`</qa_pair>` block with the missing part left empty. For example, if only questions appear:
  `<qa_pair><label>Example 1</label><question>...</question><answer></answer><solution></solution></qa_pair>`
- If multiple questions and solutions appear, wrap each question/solution pair in its own `<qa_pair>`...`</qa_pair>` block.
- If you do not see the full solution, only extract the short answer and leave the solution empty. YOU MUST KEEP QUESTIONS WITH ONLY SHORT ANSWERS !!!
** About chapter/section titles **
- Always enclose qa pairs in a `<chapter>`...`</chapter>` block, where <title>MAIN_TITLE</title> is the chapter title or section title.
- Normally, chapter/section titles appear before the questions/answers in an independent json item.
- There could be multiple `<chapter>`...`</chapter>` blocks if multiple chapters/sections exist. 
- **Any titles followed by a question/answer whose label/number is not 1, or with a score, should NOT be extracted.**
- Do not use nested titles.
- Leave the title blank if there is no chapter title. 
** About figures/diagrams **
- Whenever the question or answer/solution refers to a figure or diagram, record its "id" in question/answer/solution just like other text content.
- You MUST include all images referenced in the question/answer/solution.


If no qualifying content is found, output:
<empty></empty>

Output format (all tags run together, no extra whitespace or newlines except between entries):
<chapter><title>MAIN_TITLE_ID</title>
<qa_pair><label>...</label><question>QUESTION_IDS</question>
<answer>ANSWER(EXTRACTED FROM SOLUTION)</answer><solution>SOLUTION_IDS</solution></qa_pair>
<qa_pair><label>...</label><question>QUESTION_IDS</question>
<answer>ANSWER(EXTRACTED FROM SOLUTION)</answer><solution></solution></qa_pair>
</chapter>
<chapter><title>MAIN_TITLE</title>
<qa_pair><label>...</label><question>QUESTION_IDS</question>
<answer>ANSWER(EXTRACTED FROM SOLUTION)</answer><solution>SOLUTION_IDS</solution></qa_pair>
</chapter>


Example:
<chapter><title>1</title>
<qa_pair><label>Example 1</label><question>2,3</question>
<answer>4/5</answer><solution>5,6,7</solution></qa_pair>
<qa_pair><label>Example 2</label><question>8,9,10</question>
<answer>3.14</answer><solution></solution></qa_pair>
</chapter>
<chapter><title>12</title>
<qa_pair><label>Example 1</label><question>13,14</question>
<answer>2^6</answer><solution>16</solution></qa_pair>
</chapter>

Please now process the provided json and output your result.
\end{lstlisting}
\end{promptbox}

\subsection{Sub-question Decoupling (\Cref{sec:subquestion_decompose})}
\label{app:subquestion_decompose}
The prompt used for breaking down sub-questions in one compound question is as follows:
\begin{promptbox}[title=Sub-question Decoupling]
\begin{lstlisting}[basicstyle=\ttfamily\scriptsize\linespread{1.0}\selectfont,
                    breaklines=true,
                    breakatwhitespace=false,
                    columns=flexible,
                    xleftmargin=0pt,
                    breakindent=0pt]
You are an educational question structure analysis assistant. Below is a composite question and its corresponding answer. Please split it into several independent sub-questions.
The requirements are as follows:

1. The question may contain multiple sub-questions (e.g., (a)(b)); please accurately identify and split them one by one. Only split sub-questions with clear labels.
    Do not split implicit sub-questions (such as "What is the value of x and y?" or multiple question marks).
2. Each sub-question must be self-contained and answerable. If the original question contains contextual information, include it in each sub-question to preserve full meaning.
    If sub-questions are related (e.g., "(a) Find x. (b) Using the value of x, find y."), do not split them; keep them as one sub-question.
3. If an answer or/and solution is provided, try to match each sub-question with its corresponding part of the answer or/and solution based on semantics.
4. If the original answer or/and solution contains LaTeX formulas, preserve them exactly as they appear.
5. If the original answer or/and solution is missing or cannot be clearly aligned, leave `"sub_answer"` or/and `"sub_solution"` as an empty string.
6. The output must be a valid JSON array, where each element contains:

   * `"sub_id"`: the index of the sub-question (an integer starting from 1)
   * `"sub_question"`: the complete text of the sub-question (or "ORIGINAL" if no splitting is needed)
   * `"sub_answer"`: the corresponding answer, empty string if unavailable (or "ORIGINAL" if no splitting is needed)
   * `"sub_solution"`: the corresponding solution, empty string if unavailable (or "ORIGINAL" if no splitting is needed)
   
[Important Notice]
1. In some questions, answers or solutions, there will be figures written as `![image](image_url)`. When splitting, please keep these figure references in the corresponding sub-questions, sub-answers, or sub-solutions as EXACTLY what they are.
2. If the question does not need to be split, return an array with a single element, simplified as: [{"sub_id": 1, "sub_question": "ORIGINAL", "sub_answer": "ORIGINAL", "sub_solution": "ORIGINAL"}]
    In this case, you only need to output "ORIGINAL" instead of the full text for sub_question, sub_answer, and sub_solution, so that we can save tokens.

## Example Input:

**Question:**  
A class has 40 students, including 25 boys and 15 girls. ![image](question_images/a284h5iuh38.jpg) (a) Find the percentage of boys in the class. (b) Find the percentage of girls in the class.

**Answer:**  
(a) 62.5%. (b) 37.5%.

**Solution:**  
Percentage of boys = (25/40) * 100 = 62.5%, percentage of girls = (15/40) * 100 = 37.5%.
----------------------------------------------

## Example Output:

```json
[
  {
    "sub_id": 1,
    "sub_question": "A class has 40 students, including 25 boys and 15 girls. ![image](question_images/a284h5iuh38.jpg) Find the percentage of boys in the class.",
    "sub_answer": "62.5%.",
    "sub_solution": "Percentage of boys = (25/40) * 100 = 62.5%."
  },
  {
    "sub_id": 2,
    "sub_question": "A class has 40 students, including 25 boys and 15 girls. ![image](question_images/a284h5iuh38.jpg) Find the percentage of girls in the class.",
    "sub_answer": "37.5%.",
    "sub_solution": "Percentage of girls = (15/40) * 100 = 37.5%."
  }
]
```
Now, please split the following question according to the above requirements:
[Question]
{input_question}

[Answer]
{input_answer}

[Solution]
{input_solution}
\end{lstlisting}
\end{promptbox}

\subsection{Problem Type Classification (\Cref{sec:type_classification})}
\label{app:type_classification}
The prompt used for classifying problem types is as follows:
\begin{promptbox}[title=Type Classification]
\begin{lstlisting}[basicstyle=\ttfamily\scriptsize\linespread{1.0}\selectfont,
                    breaklines=true,
                    breakatwhitespace=false,
                    columns=flexible,
                    xleftmargin=0pt,
                    breakindent=0pt]
[Role]
You are an education expert familiar with textbook question formats at high school and university levels.
Your task is to determine the question type based on the question and answer provided.

[Possible Categories]
Choose exactly one of the following types:
1. Proof problem - requires proving a statement, identity, inequality, or property.
2. Explanation problem - asks for reasoning, causes, interpretation, principle, or conceptual explanation.
3. Fill-in problem - asks to fill in blanks, complete missing expressions, or supply intermediate steps.
4. Calculation problem - involves explicit numerical or symbolic computation, formula manipulation, or value derivation.
Even if the final answer is a short conclusion such as "thus xxx increases" or "so the velocity decreases,"
it should still be considered a Calculation problem if the majority of the reasoning is computational.
5. Multiple-choice problem - asks to choose or identify the correct option (e.g., "Which of the following").
6. Sketching/Plotting problem - requires sketching a figure, diagram, graph, or geometric representation.
7. Other - for tasks that don't fit any of the above types.

[Judgment Rules]
1. If the problem explicitly says "prove," "show that," "derive," -> classify as Proof problem.
2. If it mainly contains explanations, reasoning, or conceptual analysis without detailed calculation -> Explanation problem.
3. If the question has blanks, missing terms, or placeholders (e.g., "( )" or "____") -> Fill-in problem.
4. If there are multiple formula derivations, substitutions, or numeric results -> Calculation problem,
even if followed by a brief explanatory conclusion.
5. If it asks to select the correct answer among options (A/B/C/D, etc.) -> Multiple-choice problem.
6. If the question explicitly requires producing a figure, diagram, plot, or geometric construction -> Sketching/Plotting problem.
7. If none of these clearly apply or the problem type is mixed -> Other.

[Output Format]
Return a JSON object with the following fields:
{
  "type": "Calculation | Proof | Explanation | Fill-in | Multiple-choice | Sketching/Plotting | Other",
  "reason": "Brief justification for the classification."
}

Please determine the type of the following question and output only one of the above category names.
(Proof, Explanation, Fill-in, Calculation, Multiple-choice, Sketching, Other).

[Question]
{input_question}

[Answer]
{input_answer}
\end{lstlisting}
\end{promptbox}

\subsection{Semantic Answer Filtering (\Cref{sec:data_quality})}
\label{app:answer_extraction}
The prompt used for semantic answer extraction is as follows:
\begin{promptbox}[title=Semantic Answer Extraction]
\begin{lstlisting}[basicstyle=\ttfamily\scriptsize\linespread{1.0}\selectfont,
                    breaklines=true,
                    breakatwhitespace=false,
                    columns=flexible,
                    xleftmargin=0pt,
                    breakindent=0pt]
[Role]
You are a professional question answering system. Extract concise and accurate answers from the provided solutions. 
Output only the answer without any additional text.

Extract the answer from the following solution:

Solution: {solution}

\end{lstlisting}
\end{promptbox}

\subsection{Quality Filtering (\Cref{sec:data_quality})}
\label{app:data_quality}
The prompt used for data quality filtering is as follows:
\begin{promptbox}[title=Quality Filtering]
\begin{lstlisting}[basicstyle=\ttfamily\scriptsize\linespread{1.0}\selectfont,
                    breaklines=true,
                    breakatwhitespace=false,
                    columns=flexible,
                    xleftmargin=0pt,
                    breakindent=0pt]
[Role]
You are an education expert familiar with textbook question formats at high school and university levels.
Your task is to determine whether the provided question and answer pair is suitable to serve as a problem in an exam.

Question: {input_question}

Answer: {input_answer}

[Criteria]
1. Clarity: The question must be suitable for an exam setting, meaning it should raise **a clear problem** that requires a specific solution.
Examples, **statements without questions**, open-ended discussions and other context that do not pose a clear problem are not suitable.
Questions like "Give an example of..." that can have many valid answers are also not suitable.
You should be particularly careful with questions that **only provide a topic or theme** without a specific problem to solve.
For instance, "all primes less than 100" is not a valid question, because it does not specify what to do (listing, counting, ...) with those primes.
Instead, a question like "List all primes less than 100" or "How many primes are there less than 100?" would be suitable.
2. Relevance: The answer must directly address the question asked.
If the answer seems to be addressing a different question and is wrongly paired with the given question, it is not suitable.
3. Completeness and Self-Containment: The question and answer should be complete and self-contained, providing all necessary information for understanding and solving it without requiring external context.
Questions that rely heavily on prior context or external references are not suitable.
Answers such as "Refer to theorem X", "Corollary of previous result", "Answered in the text above", "Omitted for brevity" are not acceptable.
Incomplete questions or answers that leave out critical information are also not suitable.
4. Explicit Task Requirement: The question must contain an explicit task phrase (such as "compute", "determine", "find", "prove", "list", "show", "give the value of", etc.).
Pure expressions or noun phrases are NOT acceptable even if they are commonly understood as implicit tasks in mathematical contexts. 
If the question does not include an explicit verb specifying what the student must do, it must be judged unsuitable.
Of course, if the question is in a multiple-choice or fill-in-the-blank format, the choices or blanks themselves will serve as the explicit task requirement.
   
[Important Notice]
1. You do not need to evaluate the correctness of the answer, only whether it is appropriate and complete in relation to the question.
2. Short answer with no explanation (calculation, proof, counterexample, ...) is acceptable as long as it directly addresses the question.
3. You should be very strict in your evaluation. If any of the criteria above are not fully met, the question-answer pair should be considered unsuitable.
   
[Output Format]
Return a JSON object with the following fields:
{
"reason": "Brief justification of your judgement."
"judgement": "true | false",
} 
      
Your judgment:
\end{lstlisting}
\end{promptbox}

\section{Training and Evaluation Details}

\subsection{Detailed Training Hyperparameters}
\label{app:train_hyperparams}

\Cref{tab:training_hyperparameters_qa} lists the hyperparameters used in SFT of Qwen3-8B-Base in our experiments.

\begin{table}[ht]
\centering
\caption{Detailed Hyperparameters for SFT of Qwen3-8B-Base Instruct.}
\label{tab:training_hyperparameters_qa}
\begin{tabular}{ll}
\toprule
\textbf{HyperParameter} & \textbf{Value} \\
\midrule
Method & Full Fine-tuning \\
Learning Rate & $5 \times 10^{-5}$ \\
Learning Rate Scheduler & Cosine \\
Warmup Ratio & 0.1 \\
Training Epochs & 3.0 \\
GPUs & 8 \\
Per-GPU Batch & 2 \\
Gradient Accumulation Steps & 4 \\
Effective Batch Size & 64 \\
Optimizer & AdamW \\
Validation Split Size & 0.1 \\
Max Sequence Length & 4096 \\
Numerical Precision & BF16 \\
\bottomrule
\end{tabular}
\end{table}

\Cref{tab:training_hyperparameters} lists the hyperparameters used in SFT of Qwen3-VL-8B Instruct in our experiments.

\begin{table}[h]
\centering
\caption{Detailed Hyperparameters for SFT of Qwen3-VL-8B Instruct.}
\label{tab:training_hyperparameters}
\begin{tabular}{ll}
\toprule
\textbf{HyperParameter} & \textbf{Value} \\
\midrule
Learning Rate & $5 \times 10^{-6}$ \\
Learning Rate Scheduler & Cosine \\
Warmup Steps & 5 \\
Training Epochs & 2.0 \\
Batch Size & 128 \\
Optimizer & AdamW \\
Maximum Gradient Norm & 1.0 \\
Vision Tower & Frozen \\
Multi-modal Projector & Trainable \\
Validation Split Size & 0.05 \\
Max Sequence Length & 12,800 \\
Image Max Pixels & $589,824$ \\
Video Max Pixels & $65,536$ \\
Numerical Precision & FP16 \\
\bottomrule
\end{tabular}
\end{table}

\subsection{Answer Generation and Verification Prompts}
\label{app:gen_prompt}

The answer generation prompt used for both distillation, training and evalution is as follows:
\begin{promptbox}[title=Answer Generation]
\begin{lstlisting}[basicstyle=\ttfamily\scriptsize\linespread{1.0}\selectfont,
                    breaklines=true,
                    breakatwhitespace=false,
                    columns=flexible,
                    xleftmargin=0pt,
                    breakindent=0pt]
You are an intelligent chatbot designed for producing the answer to the given reasoning task.
Remember: DO NOT output anything else, only output the answer you generate.
Generate a solution to the given task strictly following this format:
1. Identify key components and premises of the task
2. Apply relevant principles, theorems, or methods with step-by-step derivation or argument
3. Perform any necessary calculations or logical checks with intermediate verification
4. Present the final answer or conclusion in a clear, unambiguous notation

Format Requirements:
- Prefix each step with arrow symbol
- Ensure all symbols and special characters are presented using appropriate markup (e.g., LaTeX commands for mathematical symbols, code formatting for code snippets)

Example Template:
Task: Analyze the time complexity of the following sorting algorithm and prove its correctness.

Solution:
1. Identify components:
-> Algorithm uses divide-and-conquer to split the list in half
-> Merging step compares elements pairwise

2. Apply principles:
-> Recurrence: T(n) = 2T(n/2) + O(n)
-> By Master Theorem, T(n) = O(n log n)

3. Verification:
-> Check base case T(1) = O(1)
-> Inductive step holds for n = 2^k

4. Conclusion:
-> The algorithm runs in \\boxed{O(n\\log n)} time and correctly sorts any input list.

Here is the given task you need to solve:
{question}
\end{lstlisting}
\end{promptbox}


The prompt for answer verification is as follows:
\begin{promptbox}[title=Answer Verification]
\begin{lstlisting}[basicstyle=\ttfamily\scriptsize\linespread{1.0}\selectfont,
                    breaklines=true,
                    breakatwhitespace=false,
                    columns=flexible,
                    xleftmargin=0pt,
                    breakindent=0pt]
As an answer evaluation expert, please assess whether the following answer is correct.
        
Question: {question}

Reference Answer: {reference_answer}

Current Answer: {answer}

Please carefully analyze whether the current answer is semantically consistent with the reference answer. 
Focus only on comparing the answers themselves, not on how the problem is solved.
Don't just look at the surface text, understand the essential content of the answers.
If the current answer is semantically consistent with the reference answer, even if expressed differently, it should be judged as correct.
For numerical calculation problems, also consider whether the answer is within the acceptable error range (typically 5%). Be careful to differentiate whether the question is indeed a numerical calculation or one that requires a strictly identical answer.

Please return your judgment result in JSON format:
{{"judgement_result": true}} indicates the answer is correct
{{"judgement_result": false}} indicates the answer is incorrect

Your judgment:
\end{lstlisting}
\end{promptbox}


\section{Evaluation Guidelines for Human Annotators on VQA Extraction}
\label{app:annotations}
The goal of this evaluation is to verify the structural and content integrity of VQA pairs extracted from PDFs, mainly focusing on the performance of the LLM's reconstruction logic.

\subsubsection{Precision Errors (Incorrect Extractions)}
Mark the sample as \textbf{incorrect} if any of the following extra or wrong information is present:
\begin{itemize}
    \item \textbf{QA Mismatch}: The extracted question does not correspond to the provided answer.
    \item \textbf{Hallucinations}: The text in questions or answers contains contents that do not exist in the original source document.
    \item \textbf{Ordering Issues}: QA pairs are not extracted in the proper sequence as they appear in the document.
    \item \textbf{Image Misplacement}: Images are inserted at the wrong location (e.g., an image related to Question 5 is placed inside Question 2, or an image related to Answer 2 is is placed inside Question 2).
\end{itemize}

\subsubsection{Recall Errors (Missing Information)}
Mark the sample as \textbf{incorrect} if information from the source is omitted:
\begin{itemize}
    \item \textbf{Incompleteness}: A QA pair is missing a significant portion of its required text (e.g., half of the answer is cut off), or the whole question/answer is missing.
    \item \textbf{Missing Images}: An image related to the question or answer in the PDF is not inserted into the extracted output.
\end{itemize}

\subsubsection{Acceptable Deviations}
You may mark the sample as \textbf{correct} even if the following minor issues are present:
\begin{itemize}
    \item \textbf{Text-level Parsing Glitches}: Minor character errors or formatting artifacts that do not change the fundamental meaning of the sentence. 
    \item \textbf{Formula Errors}: Complex mathematical LaTeX or symbols that are incorrectly rendered but do not affect understanding.
\end{itemize}

\section{Benchmark Details}
\label{app:benchmarks}

\subsubsection{QA Benchmarks}

We evaluate on nine QA benchmarks covering general reasoning, mathematical reasoning, and physics reasoning.

\begin{itemize}

\item \textbf{MMLU}~\citep{hendryckstest2021mmlu}:
A multidisciplinary benchmark covering 57 subjects across STEM, humanities, and social sciences.
We use the official test set (14k questions).

\item \textbf{MMLU-Pro}~\citep{wang2024mmlu-pro}:
An enhanced version of MMLU designed to improve evaluation difficulty and discriminative power.
We use the official test set (12k questions).

\item \textbf{BBH}~\citep{suzgun2022challenging}:
BIG-Bench Hard consists of 23 challenging reasoning tasks evaluating logical reasoning,
symbolic manipulation, and complex arithmetic.
We use the full dataset (6.5k questions).

\item \textbf{SuperGPQA}~\citep{pteam2025supergpqa}:
A graduate-level knowledge reasoning benchmark with broad disciplinary coverage.
We use the official test set (26.5k questions).

\item \textbf{GSM8K}~\citep{cobbe2021gsm8k}:
A grade-school mathematical reasoning benchmark consisting of multi-step arithmetic word problems.
We use the official test set (1k questions).

\item \textbf{AIME}:
Competition-level mathematics problems from the 2024 and 2025 American Invitational Mathematics Examination.
The evaluation set contains 60 problems.

\item \textbf{MATH500}~\citep{hendrycks2021math}:
A challenging subset of the MATH dataset covering algebra, geometry, number theory,
calculus, probability, and statistics.
We use the full evaluation set (500 problems).

\item \textbf{UGPhysics}~\citep{xu2025ugphysics}:
An undergraduate-level physics QA benchmark covering mechanics, electromagnetism,
thermodynamics, optics, and modern physics.
We use the official test set (5.5k questions).

\end{itemize}

\subsubsection{VQA Benchmarks}

We further evaluate multimodal reasoning on seven VQA benchmarks, including our in-domain test split and six out-of-distribution datasets.

\begin{itemize}

\item \textbf{FlipVQA-test}:
A held-out test split of FlipVQA containing 998 problems sampled from the full dataset.

\item \textbf{MMMU}~\citep{yue2024mmmu}:
A multidisciplinary multimodal benchmark designed to evaluate college-level knowledge and reasoning.
We use the validation set (900 problems).

\item \textbf{OlympicArena}~\citep{huang2024olympicarena}:
An Olympiad-level multimodal benchmark covering seven disciplines.
We use the validation split and retain only VQA problems, resulting in 268 evaluation instances.

\item \textbf{MathVerse}~\citep{zhang2024mathverse}:
A visual mathematics benchmark containing problems with varying degrees of visual information.
We use the \textit{testmini} split excluding text-only instances (3.94k problems).

\item \textbf{MathVista}~\citep{lumathvista}:
A benchmark for mathematical reasoning in visual contexts, including plots, diagrams,
and scientific figures.
We use the \textit{testmini} split (1k problems).

\item \textbf{OlympiadBench}~\citep{he2024olympiadbench}:
An Olympiad-level bilingual multimodal benchmark covering mathematics and physics.
We exclude text-only and proof-based problems, resulting in 3.68k evaluation instances.

\item \textbf{MatSciBench}~\citep{zhang2025matscibench}:
A multimodal benchmark for materials science question answering.
We use the test split excluding text-only problems (315 instances).

\end{itemize}

\section{Observations}
\subsection{Difficulty-induced Token Expansion}
\begin{figure*}
\centering
\includegraphics[width=0.5\linewidth]{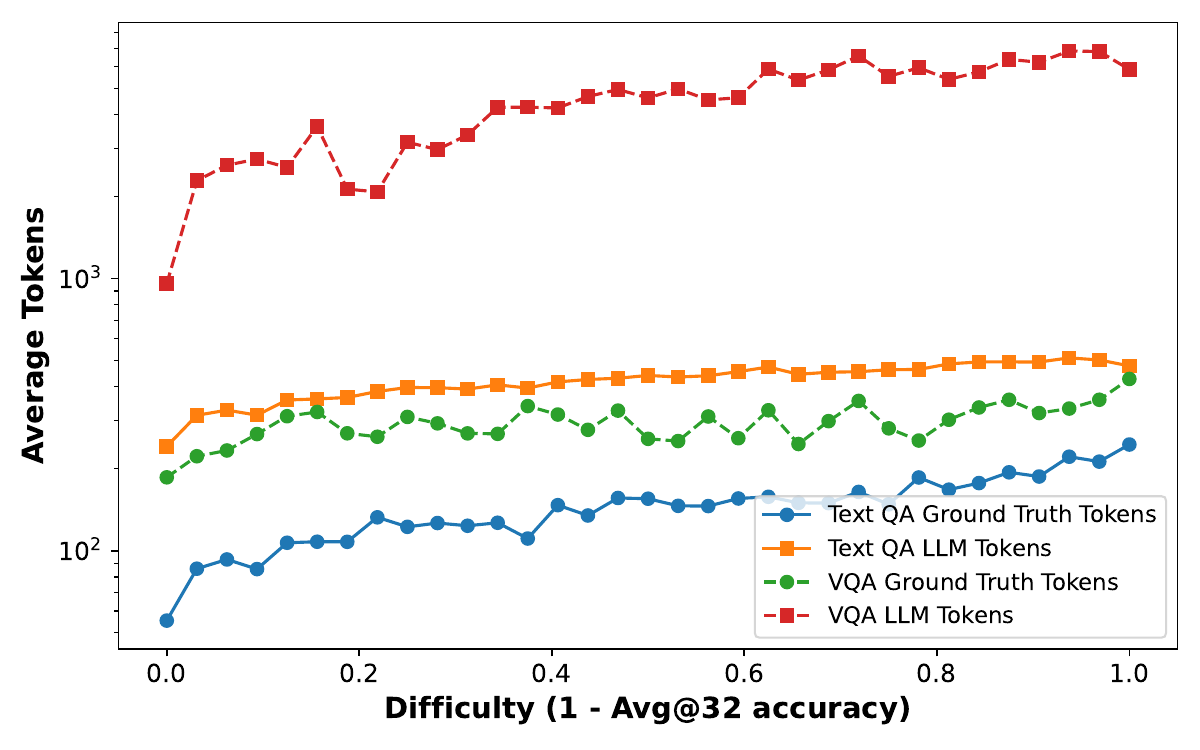}
\caption{Average tokens generated for QA and VQA tasks. Visual Question Answering (VQA) requires significantly more tokens from the LLM (red line) compared to both its ground truth and text-only tasks, especially as task difficulty increases.}
\label{fig:length}
\end{figure*}
We observe a distinct positive correlation between task difficulty and output length (as shown in Fig. ~\ref{fig:length}). Notably, this "token expansion" is disproportionately magnified in VQA, where responses often exceed 6,000 tokens—an order of magnitude longer than in QA. This suggests that multimodal reasoning requires extensive "perceptual-to-logical" translation, where exhaustive visual grounding must precede logical inference. Consequently, long-form CoT emerges as a structural hallmark of complex multimodal tasks.

\subsection{Impact of Training on High-Difficulty Samples}
Given that difficulty scales with reasoning length, we evaluate fine-tuning on high-difficulty subsets ($Difficulty \ge 0.4$). As shown in \Cref{tab:vqa_difficulty}, prioritizing these samples enhances performance on intensive reasoning benchmarks such as OlympicArena and OlympiadBench. This suggests that high-difficulty data provides the gradient density essential for deep reasoning optimization, despite a slight trade-off in general knowledge breadth on broader benchmarks like MMMU.
\begin{table*}[htbp]
  \centering
  \caption{Performance evaluation of models fine-tuned on the \textit{Full} FlipVQA dataset and its difficult subset across various reasoning benchmarks. \textit{Difficulty$\ge$ 0.4} subset keeps the problems on which the accuracy of Qwen3-VL-8B-Instruct is less than 0.6 after rolling out 32 times. }
  \label{tab:vqa_difficulty}
    \resizebox{\textwidth}{!}{
  \begin{tabular}{lccccccc}
    \toprule
    \textbf{Benchmark} & Flip-VQA Test & MMMU & OlympicArena & MathVerse & MathVista & OlympiadBench & MatSciBench \\
    \midrule
    Full Dataset &32.97 & \textbf{66.22} & 47.39 & \textbf{71.37} & 70.10 & 64.14 & \textbf{35.62} \\
    Difficulty$\ge$ 0.4 & \textbf{33.77} & 65.33 & \textbf{49.25} & 71.19 & \textbf{71.00} & \textbf{66.42} & 33.56 \\
    \bottomrule
  \end{tabular}
  }
\end{table*}

\subsection{Effect of CoT Length in Training Data}
To investigate the impact of CoT length, we selected samples from the mathematics CoT dataset whose lengths are above the median and used them for fine-tuning. As shown in Table~\ref{tab:qa_table_length}, the results indicate that training with longer CoT data improves the model’s reasoning ability, particularly on more challenging benchmarks such as AIME2024 and AIME2025.
\begin{table}[htbp]
  \centering
  \caption{Performance of Qwen3-8B-Base fine-tuned with data of different CoT lengths on several math benchmarks. $^\dagger$ Long CoT data are the samples from the math CoT dataset whose lengths are above the median.}
  \label{tab:qa_table_length}
  
  \begin{tabular}{lccccc}
    \toprule
    \textbf{Benchmark} & GSM8K & MATH500 & AIME2024 & AMIE2025 & AMIE2026\\
    \midrule
    Math Data (All) & \textbf{93.71} & 82.80 & 18.44 & 16.98 & 14.90\\
    Math Data (Long CoT $^\dagger$) & 93.40 & \textbf{84.80} & \textbf{22.71} & \textbf{20.52} & \textbf{21.56}\\
    \bottomrule
  \end{tabular}
\end{table}

\subsection{Impact of Modality Mixing}
We also study the impact of mixing QA and VQA data during training. To avoid imbalance in the data distribution, we sample a subset of the QA dataset that matches the size of the VQA dataset. We evaluate five data composition strategies: (1) VQA-only; (2) QA-only; (3) Sequential (VQA $\rightarrow$ QA); (4) Sequential (QA $\rightarrow$ VQA); and (5) Joint Training (VQA + QA). The results in \Cref{tab:vqa_modality_mix} reveal several key observations:
\begin{enumerate}
    \item \textbf{Cross-modal transfer}. The \textbf{QA-only} setting yields performance gains over the baseline on several VQA benchmarks, suggesting that improvements in textual reasoning can partially transfer to visual reasoning tasks.
    \item \textbf{Catastrophic forgetting under sequential training}. In sequential training regimes, the \textbf{QA $\rightarrow$ VQA} strategy underperforms the VQA-only setting, while the \textbf{VQA $\rightarrow$ QA} sequence nearly regresses to the QA-only baseline. This behavior indicates substantial forgetting of visual understanding when the model is subsequently fine-tuned on purely textual data.
    \item \textbf{Joint training synergy}: The joint modality setting (VQA + QA) consistently achieves the best performance, demonstrating that textual reasoning improvements can complement visual understanding when both modalities are trained jointly.
\end{enumerate}
\begin{table*}[htbp]
  \centering
  \caption{Performance across different modality data mixture methods. We explore the effects of five distinct data composition strategies. Bold and underlined values indicate the best and second-best results, respectively.}
  \label{tab:vqa_modality_mix}
   \resizebox{\textwidth}{!}{
  \begin{tabular}{lccccccc}
    \toprule
    \textbf{Benchmark} & Flip-VQA Test & MMMU & OlympicArena & MathVerse & MathVista & OlympiadBench & MatSciBench \\
    \midrule
    Baseline & 24.45 & 65.78 & 42.54 & 66.73 & \uline{70.30} & 57.35 & 29.79 \\
    VQA-only & \uline{32.97} & \uline{66.22} & \textbf{47.39} & \uline{71.37} & 70.10 & 64.14 & \uline{35.62} \\
    QA-only & 28.36 & 63.11 & 45.90 & 68.88 & 68.40 & 62.92 & 34.25 \\
    VQA $\rightarrow$ QA & 30.66 & 64.89 & 45.52 & 68.76 & 68.90 & 62.70 & 34.25 \\
    QA $\rightarrow$ VQA & 32.46 & 65.44 & 46.27 & 70.71 & \textbf{72.40} & \textbf{65.66} & 33.56 \\
    VQA + QA & \textbf{33.57} & \textbf{66.44} & \uline{47.01} & \textbf{72.34} & 70.10 & \uline{64.42} & \textbf{35.96} \\
    \bottomrule
  \end{tabular}
  }
\end{table*}

\subsection{Subject-Specific and Cross-Discipline Generalization}
We evaluate the impact of domain-specific versus joint fine-tuning across math and physics subsets (\Cref{tab:qa_table_mix}). Our analysis yields two core insights:
\begin{itemize}
    \item \textbf{Cross-domain generalization.} Models trained on physics data demonstrate competitive performance on math benchmarks, suggesting that reasoning patterns learned in physics effectively transfer to math tasks.
    \item \textbf{Cross-discipline synergy:} Joint fine-tuning on both math and physics data consistently outperforms domain-specific training, indicating that exposure to diverse CoT patterns fosters better generalization and mitigates overfitting.
\end{itemize}
\begin{table}[htbp]
  \centering
  \caption{Performance of Qwen3-8B-Base fine-tuned with data from different disciplines on several mathematical benchmarks.}
  \label{tab:qa_table_mix}
  \begin{tabular}{lccccc}
    \toprule
    \textbf{Benchmark} & GSM8K & MATH500 & AIME2024 & AIME2025 & AIME2026\\
    \midrule
    Math & 93.71 & 82.80 & 18.44 & 16.98 & 14.90 \\
    Physics & 93.55 & 82.20 & 20.73 & 16.88 & 13.02 \\
    Math + Physics & \textbf{93.86} & \textbf{84.20} & \textbf{21.56} & \textbf{22.71} & \textbf{17.08} \\
    \bottomrule
  \end{tabular}
\end{table}

\subsection{Difficulty Distribution}
\begin{figure}[h]
    \centering
    \begin{subfigure}[b]{0.48\linewidth}
        \centering
        \includegraphics[width=\linewidth]{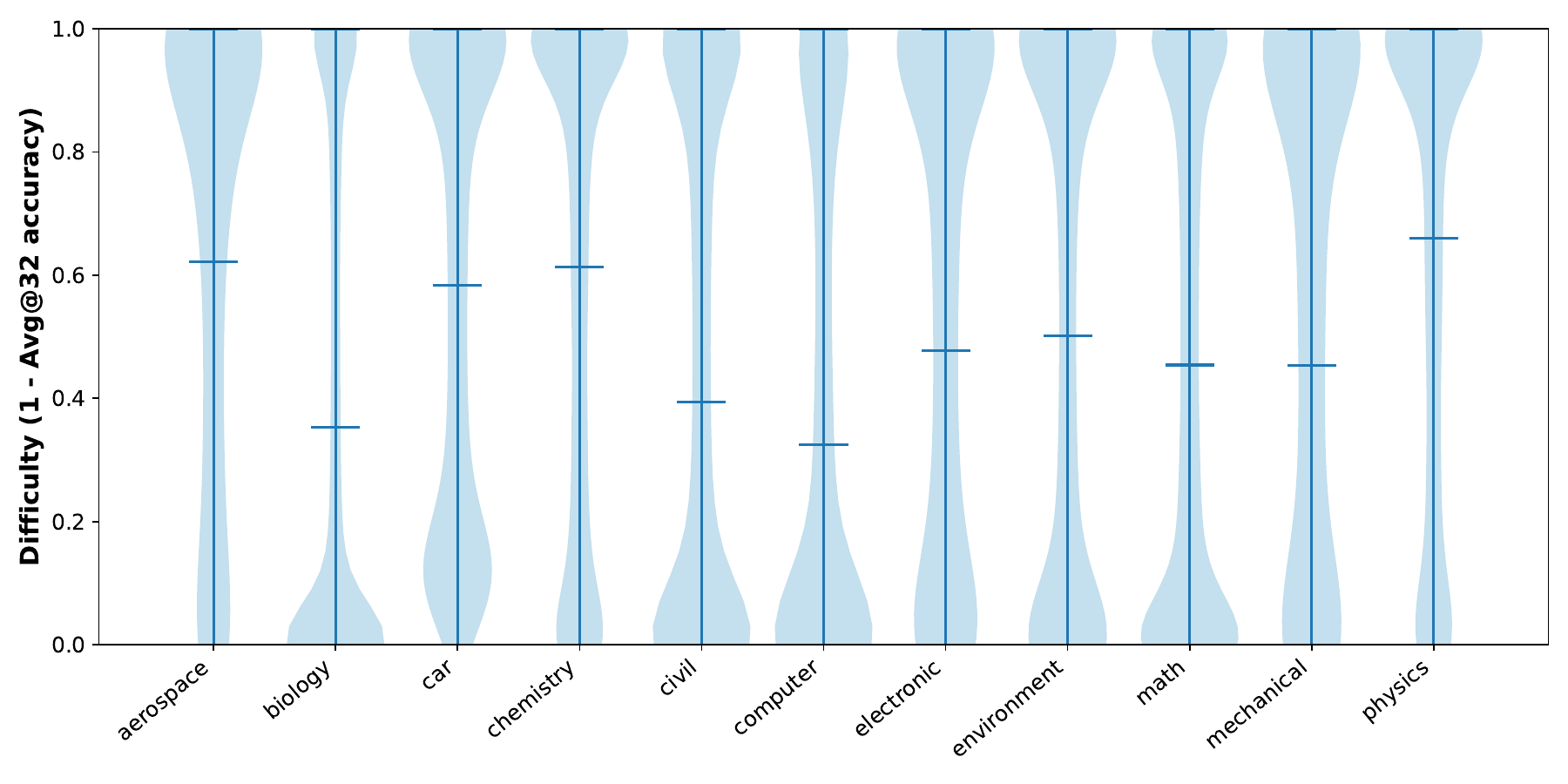}
        \caption{Text-only QA} 
        \label{fig:violin_qa}
    \end{subfigure}
    \hfill 
    \begin{subfigure}[b]{0.48\linewidth}
        \centering
        \includegraphics[width=\linewidth]{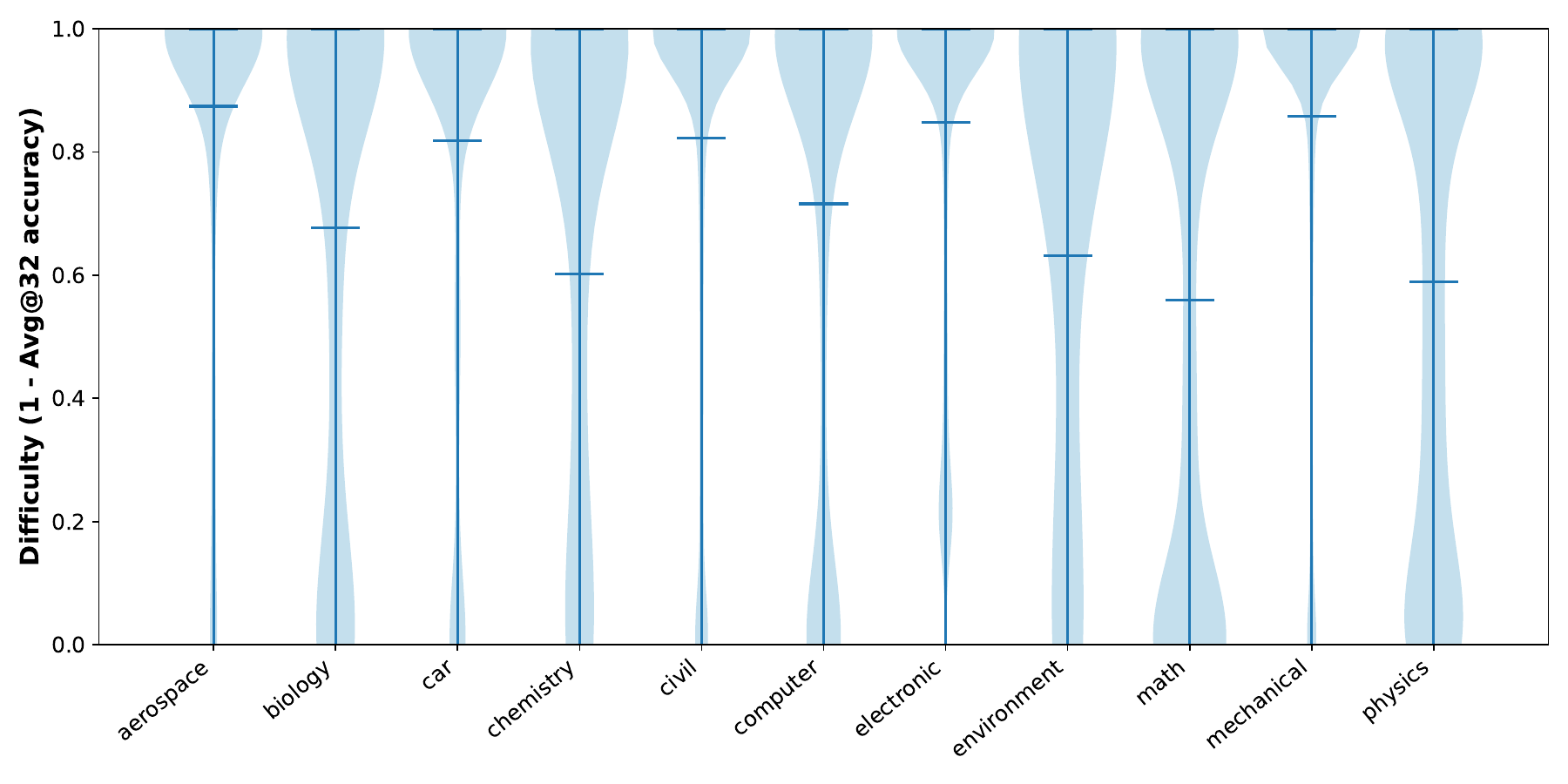}
        \caption{VQA pairs}
        \label{fig:violin_vqa}
    \end{subfigure}
    
    \caption{Difficulty distributions across different subjects for text-only QA (left) and VQA pairs (right) in our FlipVQA dataset. Each violin plot illustrates the kernel density estimation of difficulty scores, revealing a consistent bimodal pattern across disciplines.}
    \label{fig:violin}
\end{figure}

We evaluate question difficulty based on model success rates across multiple rollouts. As shown in Fig. \ref{fig:violin}, the difficulty exhibits a pronounced bimodal distribution: questions tend to cluster at the extremes (0 or 1), indicating that the model either masters the underlying concept or fails consistently. This "all-or-nothing" pattern is consistent across all subjects. Furthermore, Fig. \ref{fig:violin} reveals that domain-specific engineering (e.g., Aerospace) and hard sciences (e.g., Physics) present higher challenges compared to Mathematics. Notably, VQA pairs consistently yield higher difficulty scores than their QA counterparts, highlighting the added complexity of cross-modal reasoning.

\end{document}